\documentclass[runningheads]{llncs}
\usepackage{graphicx}
\usepackage[misc,geometry]{ifsym}
\usepackage{epsfig}
\usepackage{amsmath}
\usepackage{amssymb}
\usepackage{multirow}
\usepackage{floatrow}
\usepackage{adjustbox}
\usepackage{tabularx}
\usepackage{booktabs}
\usepackage{flushend}
\usepackage[table, dvipsnames]{xcolor}
\usepackage{caption}
\usepackage{subcaption}
\usepackage{comment}
\usepackage[colorlinks=true, linkcolor=blue, urlcolor=blue, citecolor = green]{hyperref}%
\usepackage{floatrow}
\begin{document}
\title{Transfer Learning for Scene Text Recognition in Indian Languages}
%
%
\author{Sanjana Gunna (\Letter)\orcidID{0000-0003-3332-8355}, Rohit Saluja\orcidID{0000-0002-0773-3480},  \and \\ C. V. Jawahar\orcidID{0000-0001-6767-7057}}

%
\authorrunning{Gunna et al.}
%
\institute{
Centre for Vision Information Technology \\
International Institute of Information Technology, Hyderabad - 500032, INDIA \\
\url{https://github.com/firesans/STRforIndicLanguages} \\
\email{\{sanjana.gunna,rohit.saluja\}@research.iiit.ac.in}, 
\email{jawahar@iiit.ac.in}
}
%
\maketitle              
\begin{abstract}
Scene text recognition in low-resource Indian languages is challenging because of complexities like multiple scripts, fonts, text size, and orientations. In this work, we investigate the power of transfer learning for all the layers of deep scene text recognition networks from English to two common Indian languages. We perform experiments on the conventional CRNN model and STAR-Net to ensure generalisability. To study the effect of change in different scripts, we initially run our experiments on synthetic word images rendered using Unicode fonts. We show that the transfer of English models to simple synthetic datasets of Indian languages is not practical. Instead, we propose to apply transfer learning techniques among Indian languages due to similarity in their n-gram distributions and visual features like the vowels and conjunct characters. We then study the transfer learning among six Indian languages with varying complexities in fonts and word length statistics. We also demonstrate that the learned features of the models transferred from other Indian languages are visually closer (and sometimes even better) to the individual model features than those transferred from English. We finally set new benchmarks for scene-text recognition on Hindi, Telugu, and Malayalam datasets from IIIT-ILST and Bangla dataset from MLT-17 by achieving 6\%, 5\%, 2\%, and 23\% gains in Word Recognition Rates (WRRs) compared to previous works. We further improve the MLT-17 Bangla results by plugging in a novel correction BiLSTM into our model. We additionally release a dataset of around 440 scene images containing 500 Gujarati and 2535 Tamil words. WRRs improve over the baselines by 8\%, 4\%, 5\%, and 3\% on the MLT-19 Hindi and Bangla datasets and the Gujarati and Tamil datasets. 
\end{abstract}
\begin{figure}[ht]
    \centering
    \includegraphics[width=0.32\linewidth]{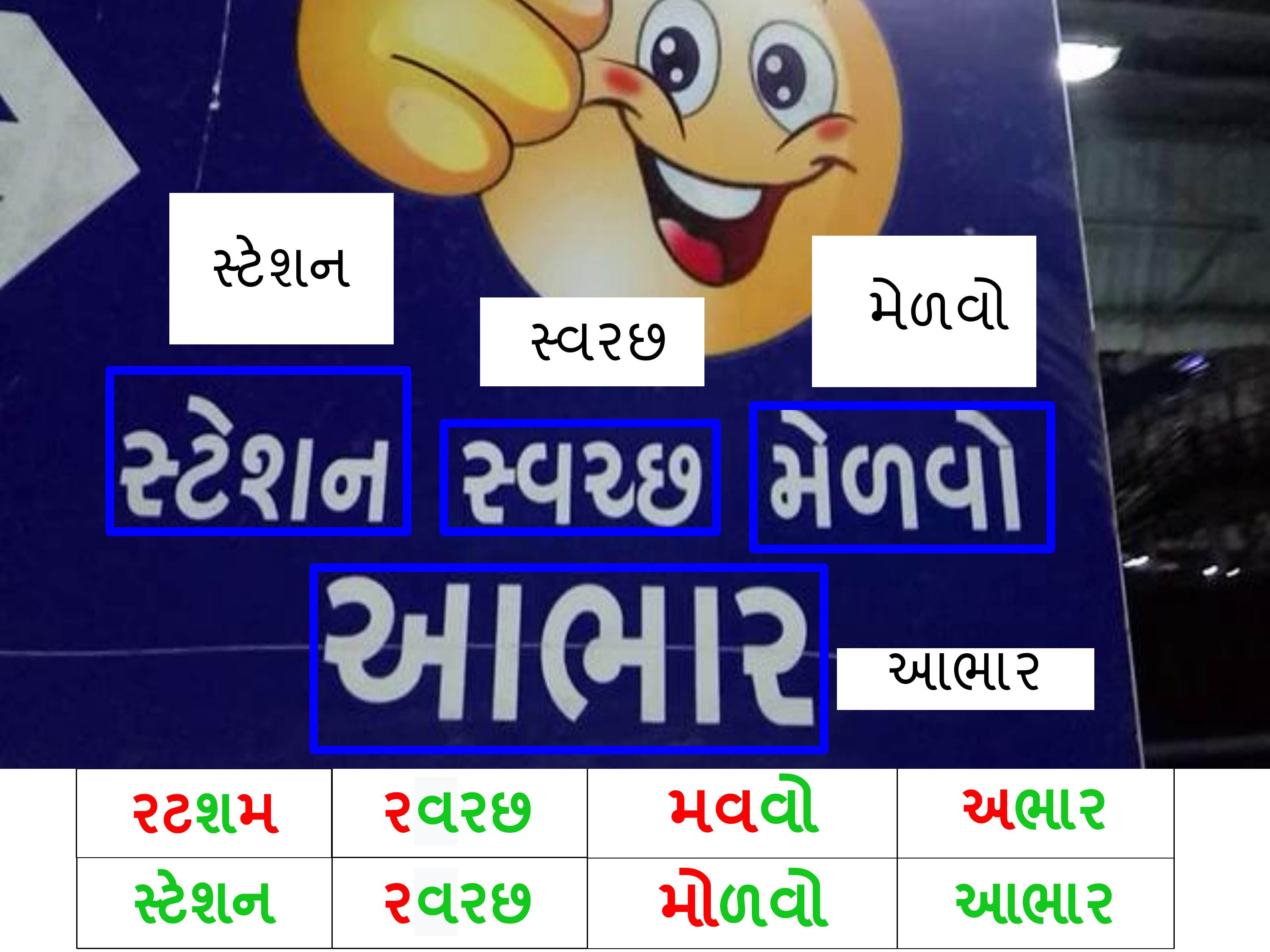}
    \centering
    \includegraphics[width=0.32\linewidth]{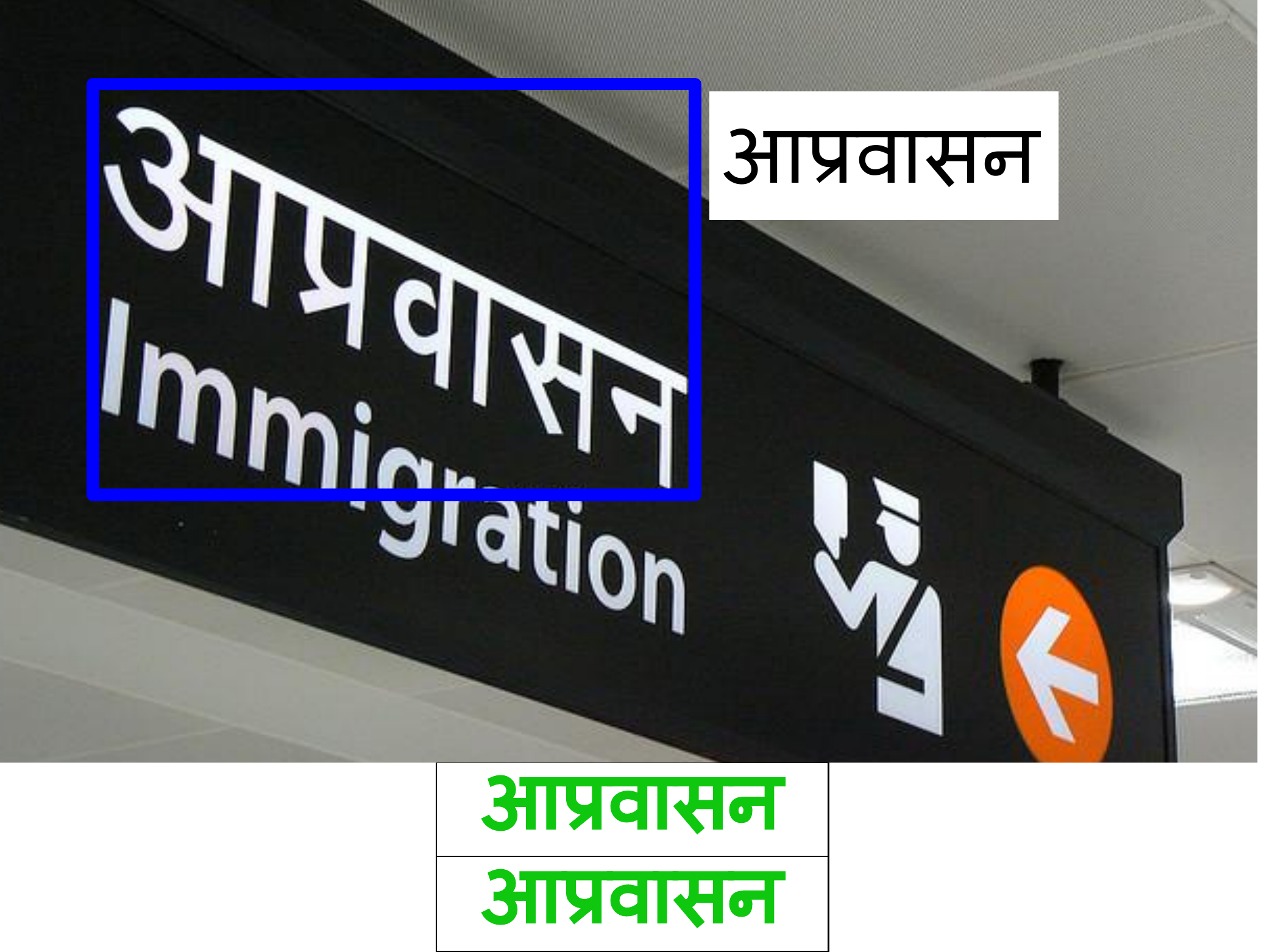}
    \centering
    \includegraphics[width=0.32\linewidth]{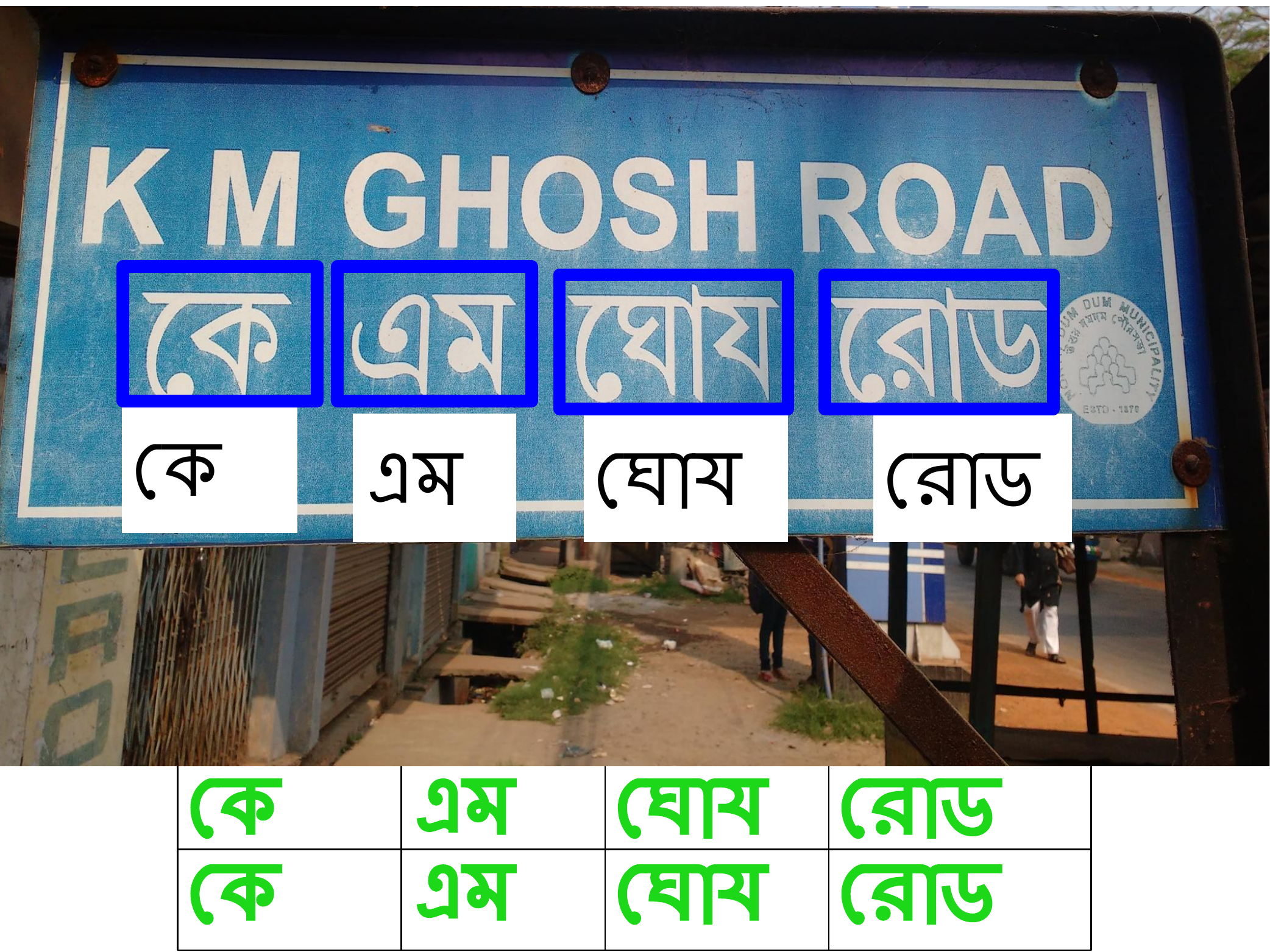}
    \\
    \centering
    \includegraphics[width=0.32\linewidth]{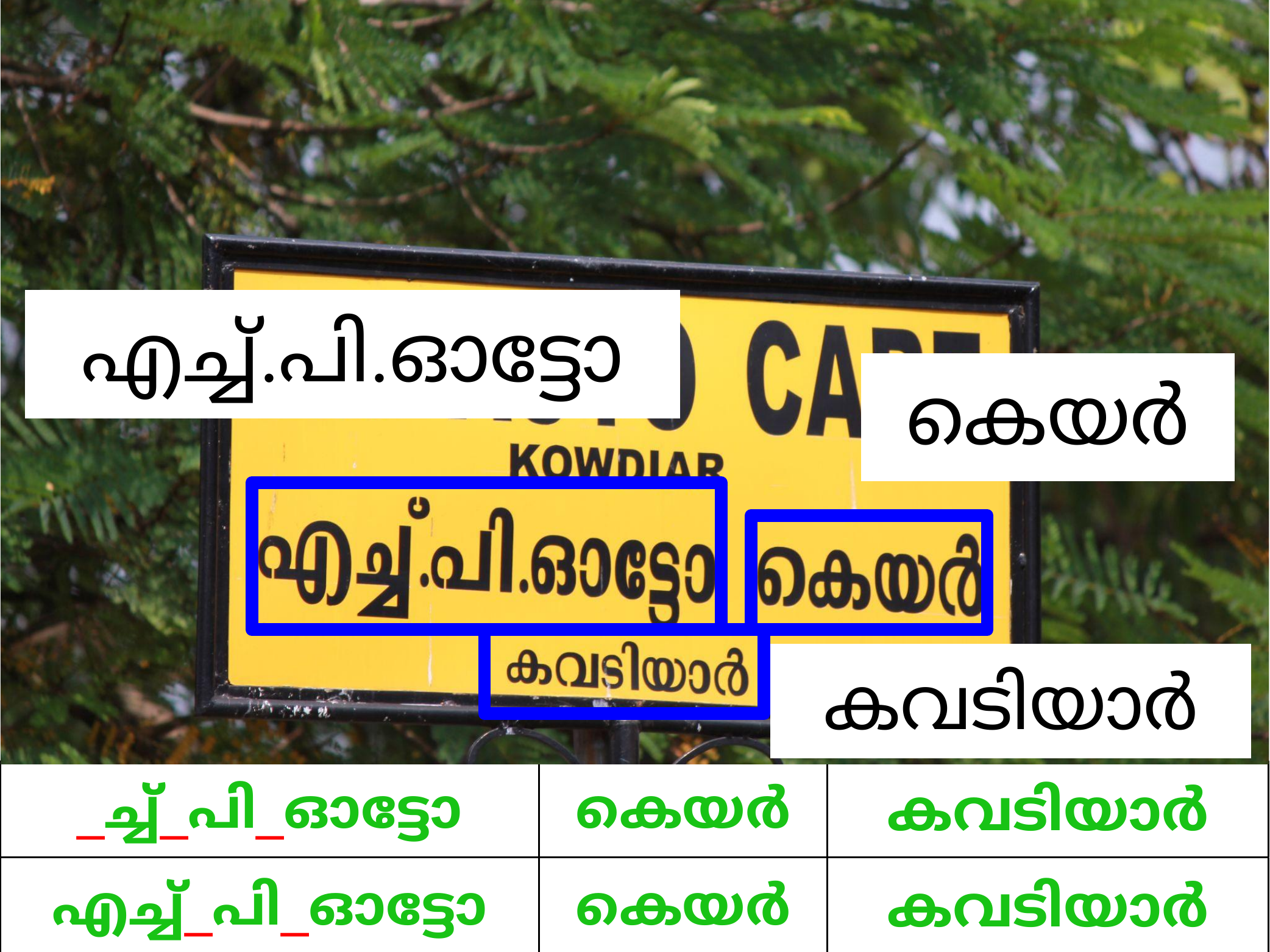}
    \centering
    \includegraphics[width=0.32\linewidth]{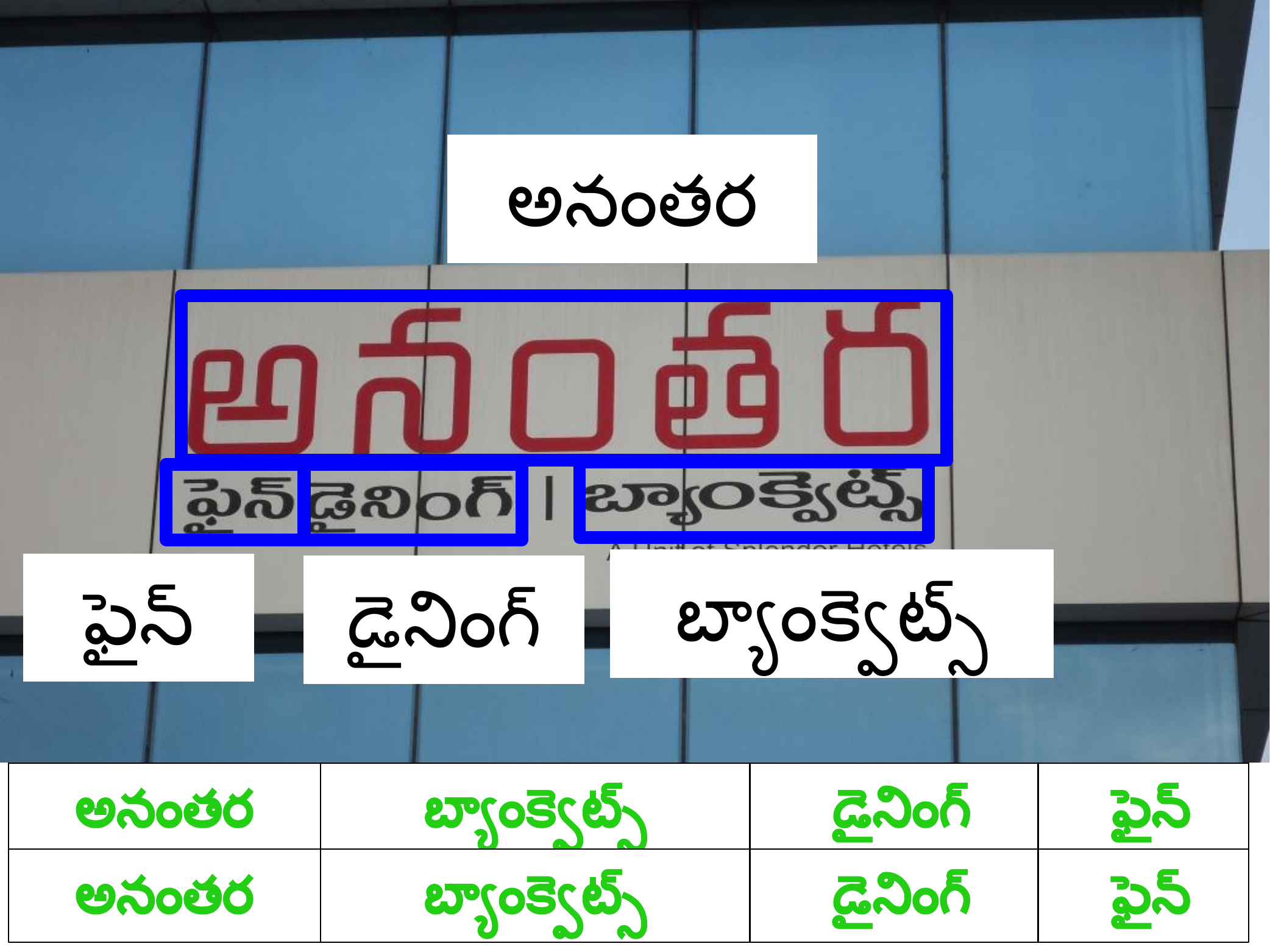}
    \centering
    \includegraphics[width=0.32\linewidth]{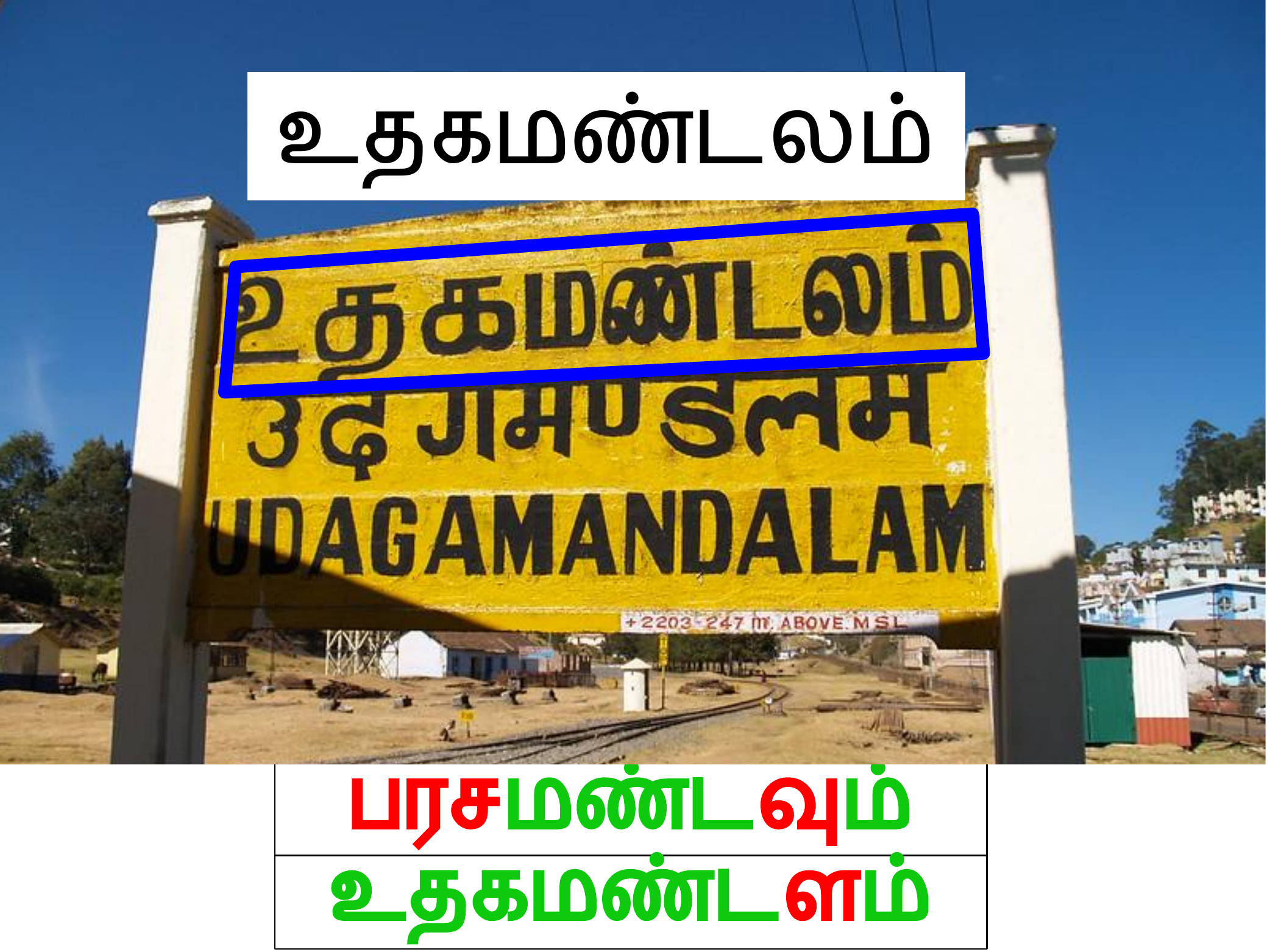}
    \caption{Clockwise from top-left; ``{\bf Top:} Annotated Scene-text images, {\bf Bottom:} Baselines' predictions (row-1) and Transfer Learning models' predictions (row-2)", from Gujarati, Hindi, Bangla, Tamil, Telugu and Malayalam. Green, red, and ``\_" represent correct predictions, errors, and missing characters, respectively. }
    \label{fig:sample_scene_images}
\end{figure}
\section{Introduction}\label{sec:Intro}
Scene-text recognition or Photo-Optical Character Recognition (Photo-OCR) aims to read scene-text in natural images. It is an essential step for a wide variety of computer vision tasks and has enjoyed significant success in several commercial applications~\cite{lee2016recursive}. Photo-OCR has diverse applications like helping the visually impaired, data mining of street-view-like images for information used in map services, and geographic information systems~\cite{buvsta2018e2e}. Scene-text recognition conventionally involves two steps; i) Text detection and ii) Text recognition. Text detection typically consists of detecting bounding boxes of word images~\cite{huang2019mask}. The text recognition stage involves reading cropped text images obtained from the text detection stage or from the bounding box annotations~\cite{mathew2017benchmarking}. In this work, we focus on the task of text recognition.

The multi-lingual text in scenes is a crucial part of human communication and globalization. Despite the popularity of recognition algorithms, non-Latin language advancements have been slow. Reading scene-text in such low resource languages is a challenging research problem as it is generally unstructured and appears in diverse conditions such as scripts, fonts, sizes, and orientations. Hence a large amount of dataset is usually required to train the scene-text recognition models. Conventionally, the synthetic dataset is used to deal with the problem since a large number of fonts are available in such low resource languages~\cite{mathew2017benchmarking}. The synthetic data may also serve as an exciting asset to perform controlled experiments, e.g., to study the effect of transfer learning with the change in script or language text. We investigate such effects for transfer from English to two Indian languages in this work, i.e., Hindi and Gujarati. We also explore the transferability of features among six different Indian languages. We share $2500$ scene text word images obtained from over $440$ scenes in Gujarati and Tamil to demonstrate such effects. In Fig.~\ref{fig:sample_scene_images}, we illustrate the sample annotated images from our datasets, and IIIT-ILST and MLT datasets, and the predictions of our models. The overall methodology we follow is that we first generate the synthetic datasets in the six Indian languages. We describe the dataset generation process and motivate the work in Section~\ref{sec:datasets}. We then train the two deep neural networks we introduce in Section~\ref{sec:models} on the individual language datasets. Subsequently, we apply transfer-learning on all the layers of different networks from one language to another. Finally, as discussed in Section~\ref{sec:experiments}, we fine-tune the networks on standard datasets and examine their performance on real scene-text images in Section~\ref{sec:results}. We finally conclude the work in Section~\ref{sec:conclusions}. The summary of our contributions are as follows:
\begin{enumerate}
    \item We investigate the transfer learning of complete scene-text recognition models i) from English to two Indian languages and ii) among the six Indian languages, i.e., Gujarati, Hindi, Bangla, Telugu, Tamil, and Malayalam.
    \item We also contribute two datasets of around $500$ word images in Gujarati and $2535$ word images in Tamil from a total of $440$ Indian scenes.
    \item We achieve gains of $6\%$, $5\%$, and $2\%$ in Word Recognition Rates (WRRs) on IIIT-ILST Hindi, Telugu, and Malayalam datasets in comparison to previous works~\cite{mathew2017benchmarking,saluja2019ocrgo}. 
    On the MLT-19 Hindi and Bangla datasets and our Gujarati and Tamil datasets, we observe the WRR gains of $8\%$, $4\%$, $5\%$, and $3\%$, respectively, over our baseline models.
    \item For the MLT-17 Bangla dataset, we show a striking improvement of $15\%$ in Character Recognition Rate (CRR) and $24\%$ in WRR compared to Bušta et al.~\cite{buvsta2018e2e}, by applying transfer-learning from another Indian language and plugging in a novel correction RNN layer into our model.
\end{enumerate}
\subsection{Related Work}\label{sec:relatedwork}
We now discuss datasets and associated works in the field of photo-OCR.

{\bf Works of Photo-OCR on Latin Datasets:} As stated earlier, the process of Photo-OCR conventionally includes two steps: i) Text detection and ii) Text recognition. With the success of Convolutional Neural Networks (CNN) for object detection, the works have been extended to text detection, treating words or lines as the objects~\cite{tian2016detecting,zhou2017east,liu2018fots}. Liao et al.~\cite{liao2017textboxes} extend such works to real-time detection in scene images. Karatzas et al.~\cite{karatzas2015icdar} and  Bušta et al.~\cite{buvsta2017deep} present more efficient and accurate methods for text detection. 
Towards reading scene-text, Wang et al.~\cite{wang2011end} propose an object recognition pipeline based on a ground truth lexicon. It achieves competitive performance without the need for an explicit text detection step. Shi et al.~\cite{shi2016end} propose a  Convolutional  Recurrent  Neural  Network  (CRNN) architecture, which integrates feature extraction, sequence modeling, and transcription into a unified framework. The model achieves remarkable performances in both lexicon-free and lexicon-based scene-text recognition tasks. Liu et al.~\cite{liu2016star} introduce Spatial Attention Residue Network (STAR-Net) with spatial
transformer-based attention mechanism to remove image distortions, residue
convolutional blocks for feature extraction, and an RNN block for decoding the text. Shi et al.~\cite{shi2018aster} propose a segmentation-free Attention-based method for Text Recognition (ASTER) by adopting Thin-Plate-Spline (TPS) as a rectification unit. It tackles complex distortions and reduces the difficulty of irregular text recognition. The model incorporates ResNet to improve the network's feature representation module and employs an attention-based mechanism combined with a Recurrent Neural Network (RNN) to form the prediction module. Uber-Text is a large-scale Latin dataset that contains around $117K$ images captured from $6$ US cities~\cite{zhang2017uber}. The images are available with line-level annotations. The French Street Name Signs (FSNS) data contains around $1000K$ annotated images, each with four street sign views. Such datasets, however, contain text-centric images. Reddy et al.~\cite{reddy2020roadtext} recently release RoadText-1K to introduce challenges with generic driving scenarios where the images are not text-centric.  RoadText-1K includes $1000$ video clips (each $10$ seconds long at $30$ fps) from the BDD dataset, annotated with English transcriptions~\cite{yu2018bdd100k}.

{\bf Works of Photo-OCR on Non-Latin Datasets:} 
Recently, there has been an increasing interest in scene-text recognition for non-Latin languages such as Chinese, Korean, Devanagari, Japanese, etc. 
Several datasets like RCTW ($12 k$ scene images), ReCTS-25k ($25k$ signboard images), CTW ($32 k$ scene images), and RRC-LSVT ($450k$ scene images) from ICDAR'19 Robust Reading Competition (RRC) exist for Chinese~\cite{shi2017icdar2017,zhang2019icdar,yuan2019ctw,sun2019icdar}. Arabic datasets like ARASTEC ($260$  images of signboards, hoardings, and  advertisements) and ALIF ($7k$ text images from TV Broadcast) also exist in the scene-text recognition community~\cite{tounsi2015arabic,yousfi2015alif}. Korean and Japanese scene-text recognition datasets include KAIST ($2,385$ images from signboards, book  covers, and English and Korean characters) and DOST ($32k$ sequential images)~\cite{jung2011touch,iwamura2016downtown}. The MLT dataset available from the ICDAR'17 RRC contains $18k$ scene images (around $1-2k$ images per language) in Arabic, Bangla, Chinese, English, French, German, Italian, Japanese, and Korean~\cite{nayef2017icdar2017}. The ICDAR'19 RRC builds MLT-19 over top of MLT-17 to contain $20k$ scene images containing text from Arabic, Bangla, Chinese, English, French, German, Italian, Japanese, Korean, and Devanagari~\cite{nayef2019icdar2019}. The RRC also provides $277k$ synthetic images in these languages to assist the training. 
Mathew et al.~\cite{mathew2017benchmarking} train the conventional encoder-decoder, where Convolutional Neural Network (CNN) encodes the word image features. An RNN decodes them to produce text on synthetic data for Indian languages. Here an additional connectionist temporal classification (CTC) layer aligns the RNN's output to labels. The work also releases an IIIT-ILST dataset for testing that reports Word Recognition Rates (WRRs) of $42.9\%$, $57.2\%$, and $73.4\%$ on $1K$ real images in Hindi, Telugu, and Malayalam, respectively. Bušta et al.~\cite{buvsta2018e2e} proposes a CNN (and CTC) based method for text localization, script identification, and text recognition. The model is trained and tested on $11$ languages of MLT-17 dataset. The WRRs are above $65\%$ for Latin and Hangul and are below $47\%$ for the remaining languages. The WRR reported for Bengali is $34.20\%$. Recently, an OCR-on-the-go model and obtain the WRR of $51.01\%$ on the IIIT-ILST Hindi dataset and the Character Recognition Rate (CRR) of $35\%$ on a multi-lingual dataset containing $1000$ videos in English, Hindi, and Marathi~\cite{saluja2019ocrgo}. Around $2322$ videos in these languages recorded with controlled camera movements like tilt, pan, etc., are additionally shared at \url{https://catalist-2021.github.io/}.

{\bf Transfer Learning in Photo-OCR:} With the advent of deep learning in the last decade, transfer learning became an essential part of vision models for tasks such as detection and segmentation.~\cite{ren2015faster,romera2017erfnet}. The CNN layers pre-trained from the Imagenet classification dataset are conventionally used in such models for better initialization and performance~\cite{russakovsky2015imagenet}. The scene-text recognition works also use the CNN layers from the models pre-trained on Imagenet dataset~\cite{shi2016end,liu2016star,shi2018aster}. However, to our best knowledge, there are no significant efforts on transfer learning from one language to another in the field of scene-text recognition, although transfer learning seems to be naturally suitable for reading low resource languages. We investigate the possibilities of transfer learning in all the layers of deep photo-OCR models. 
\begin{table}[t]
\resizebox{0.75\textwidth}{!}
{%
\centering
\begin{tabular}{lcccrcr}
\toprule
Language & \# Images & Train & Test  & ~ & $\mu$, $\sigma$ word length & \# Fonts \\
\midrule
English & 17.5M & 17M & 0.5M & ~ & 5.12, 2.99 & $>$1200\\ 
Gujarati & 2.5M & 2M & 0.5M& ~ & 5.95, 1.85 & 12\\
Hindi & 2.5M & 2M & 0.5M & ~ & 8.73, 3.10& 97\\
Bangla & 2.5M & 2M & 0.5M & ~ & 8.48, 2.98 & 68\\
Tamil & 2.5M & 2M & 0.5M & ~ & 10.92, 3.75& 158\\
Telugu & 5M & 5M & 0.5M & ~ & 9.75, 3.43 & 62\\
Malayalam & 7.5M & 7M & 0.5M & ~ & 12.29, 4.98 & 20\\
\bottomrule
\end{tabular}
}
\caption{Statistics of Synthetic Data. $\mu$, $\sigma$ represent mean, standard deviation.}
\label{tab:synth_data}
\end{table}
\section{Datasets and Motivation}\label{sec:datasets}
We now discuss the datasets we use and the motivation for our work.

{\bf Synthetic Datasets:} As shown in Table~\ref{tab:synth_data}, we generate $2.5M$, or more, word images each in Hindi, Bangla, Tamil, Telugu, and Malayalam\footnote{For Telugu and Malayalam, our models trained on $2.5M$ word images achieved results lower than previous works, so we generate more examples equal to Mathew et al.~\cite{mathew2017benchmarking}.} 
with the methodology proposed by Mathew et al.~\cite{mathew2017benchmarking}. For each Indian language, we use $2M$ images for training our models and the remaining set for testing. 
\begin{figure}[t]
    \centering
    \includegraphics[trim=0 150 0 0, clip,width=\linewidth]{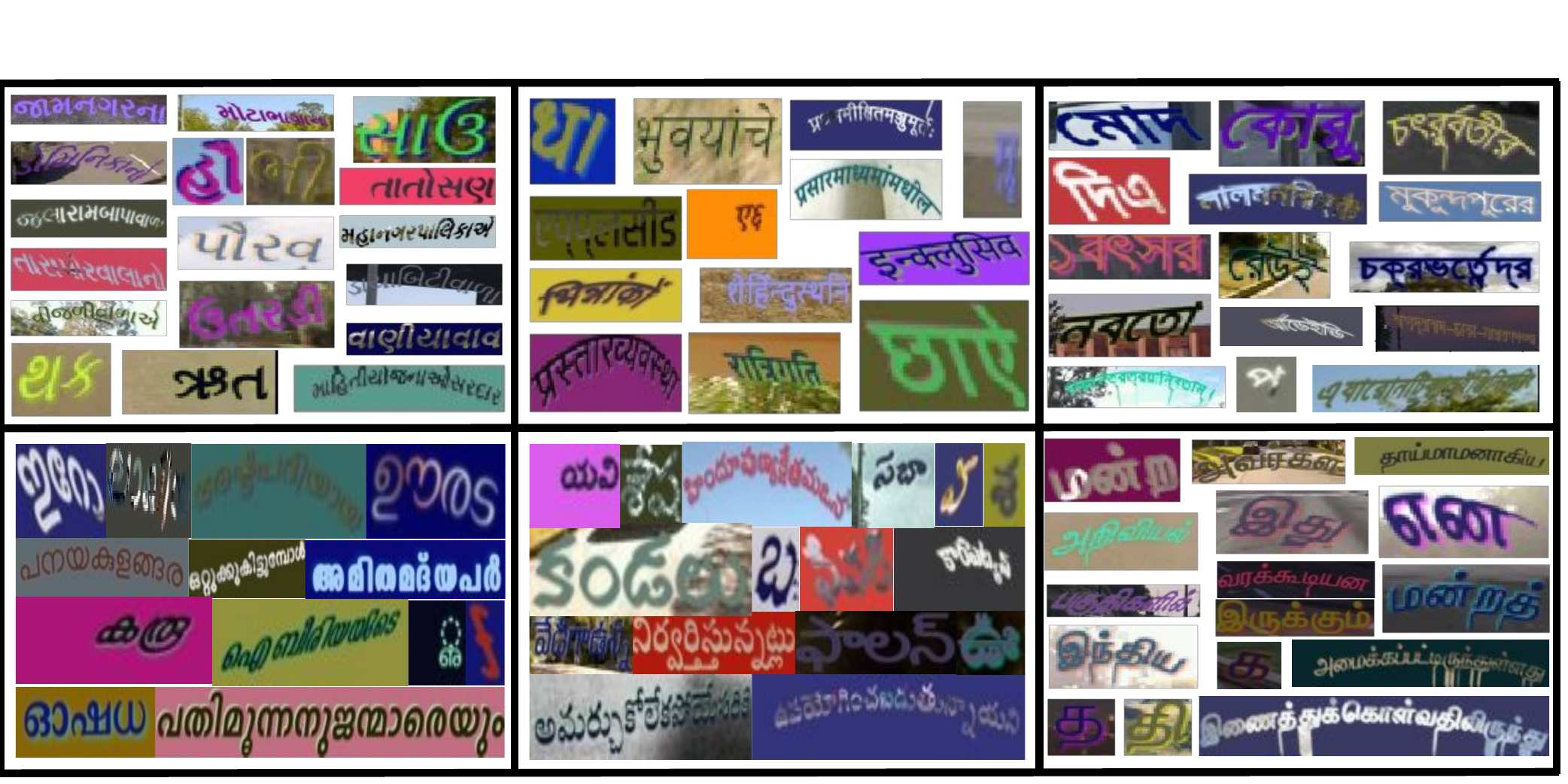}%
    \caption{Clockwise from top-left: synthetic word images in Gujarati, Hindi, Bangla, Tamil, Telugu, \& Malayalam. Notice that a top-connector line connects the characters to form a word in Hindi or Bangla. Some vowels and characters appear above and below the generic characters in Indian languages, unlike English.}
    \label{fig:sample_synth_images}
\end{figure}
\begin{figure}[h!t]
    \centering
    \begin{subfigure}[b]{0.55\textwidth}
    \centering
    \includegraphics[trim=0 35 0 110,clip,width=1\linewidth]{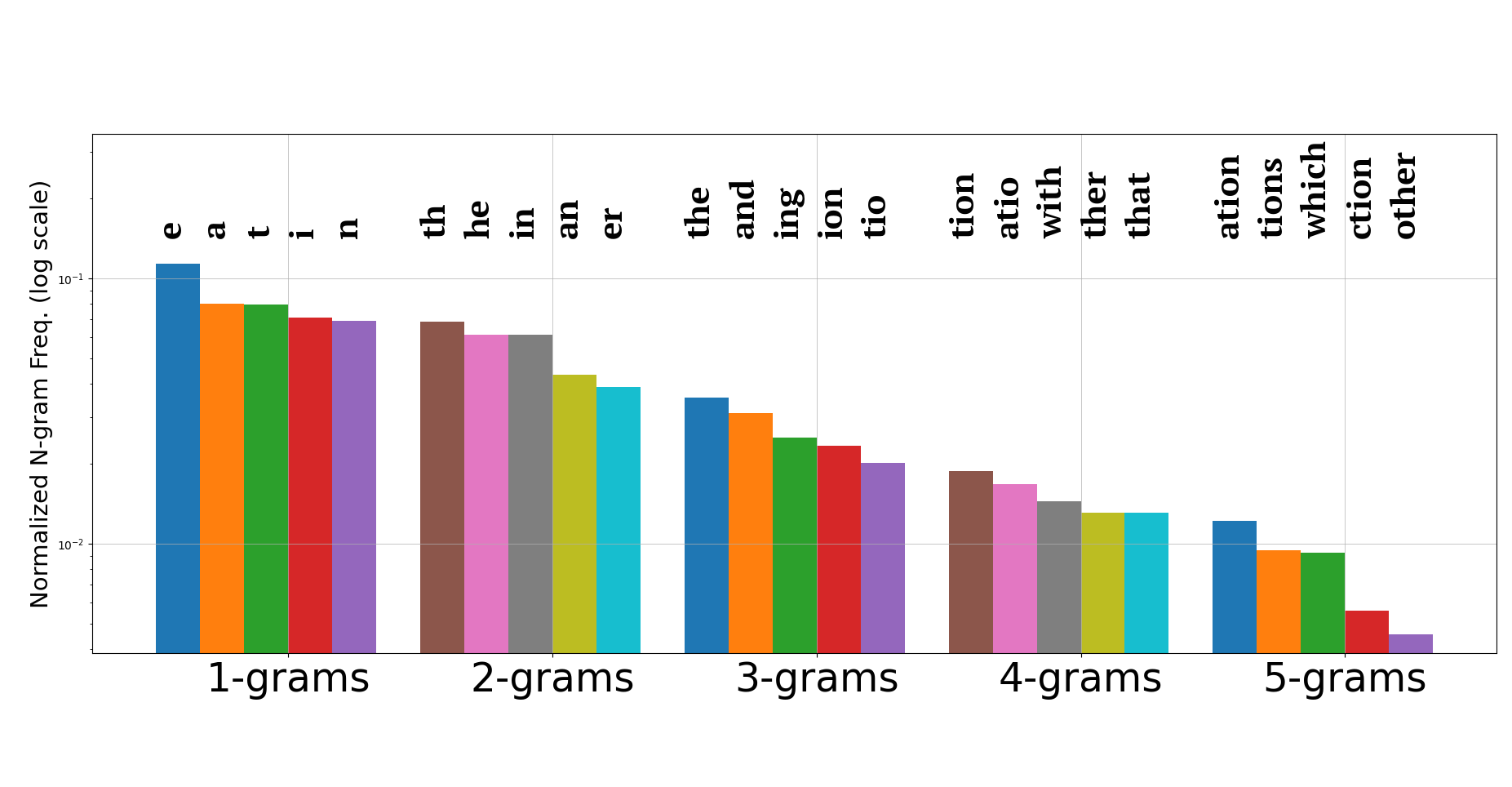}
    \end{subfigure}
    \begin{subfigure}[b]{0.435\textwidth}
    \centering
    \includegraphics[width=1\linewidth]{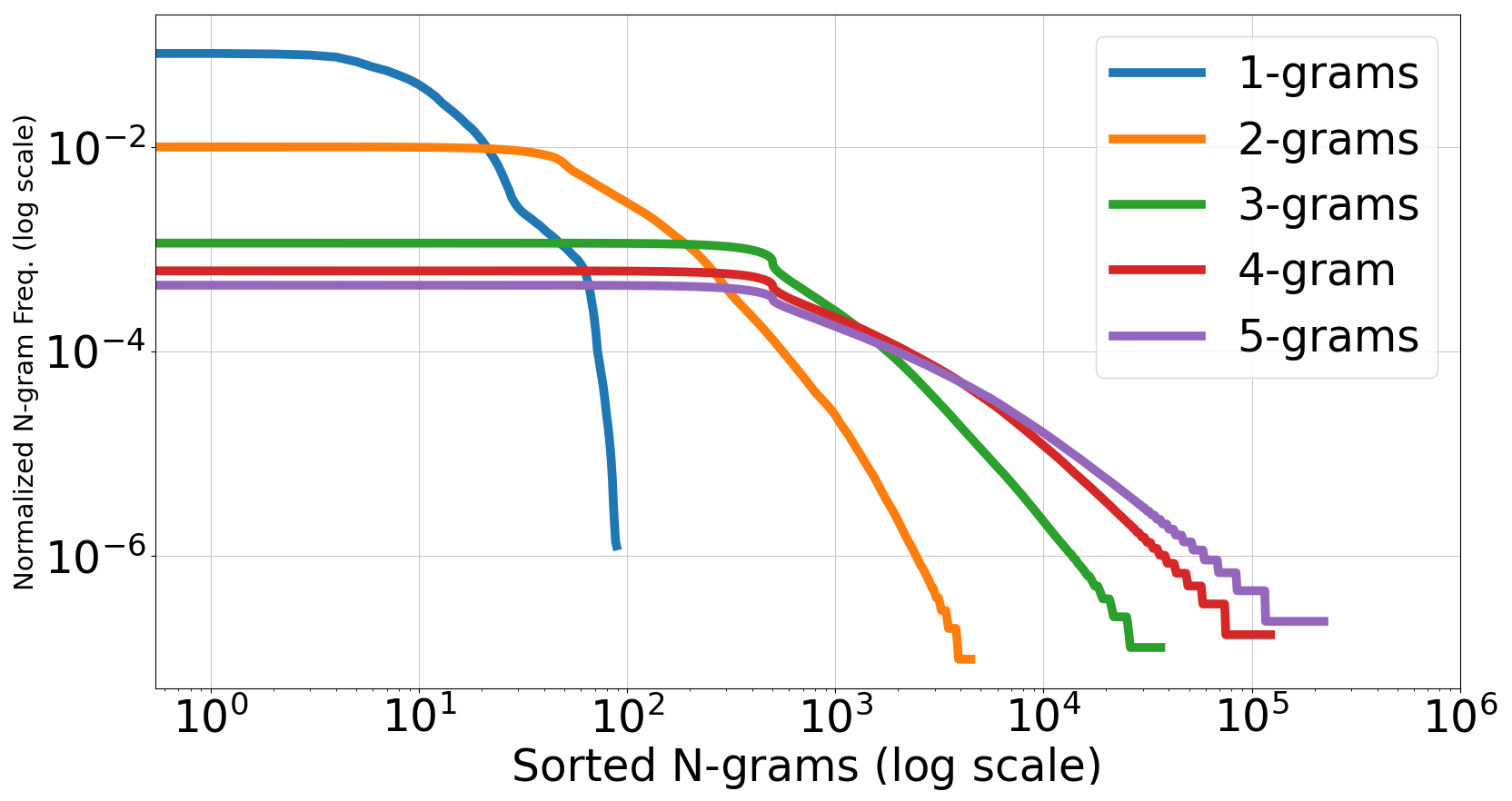}%
    \end{subfigure}
    \begin{subfigure}[b]{0.55\textwidth}
    \centering
    \includegraphics[trim=0 35 0 110, clip,width=1\linewidth]{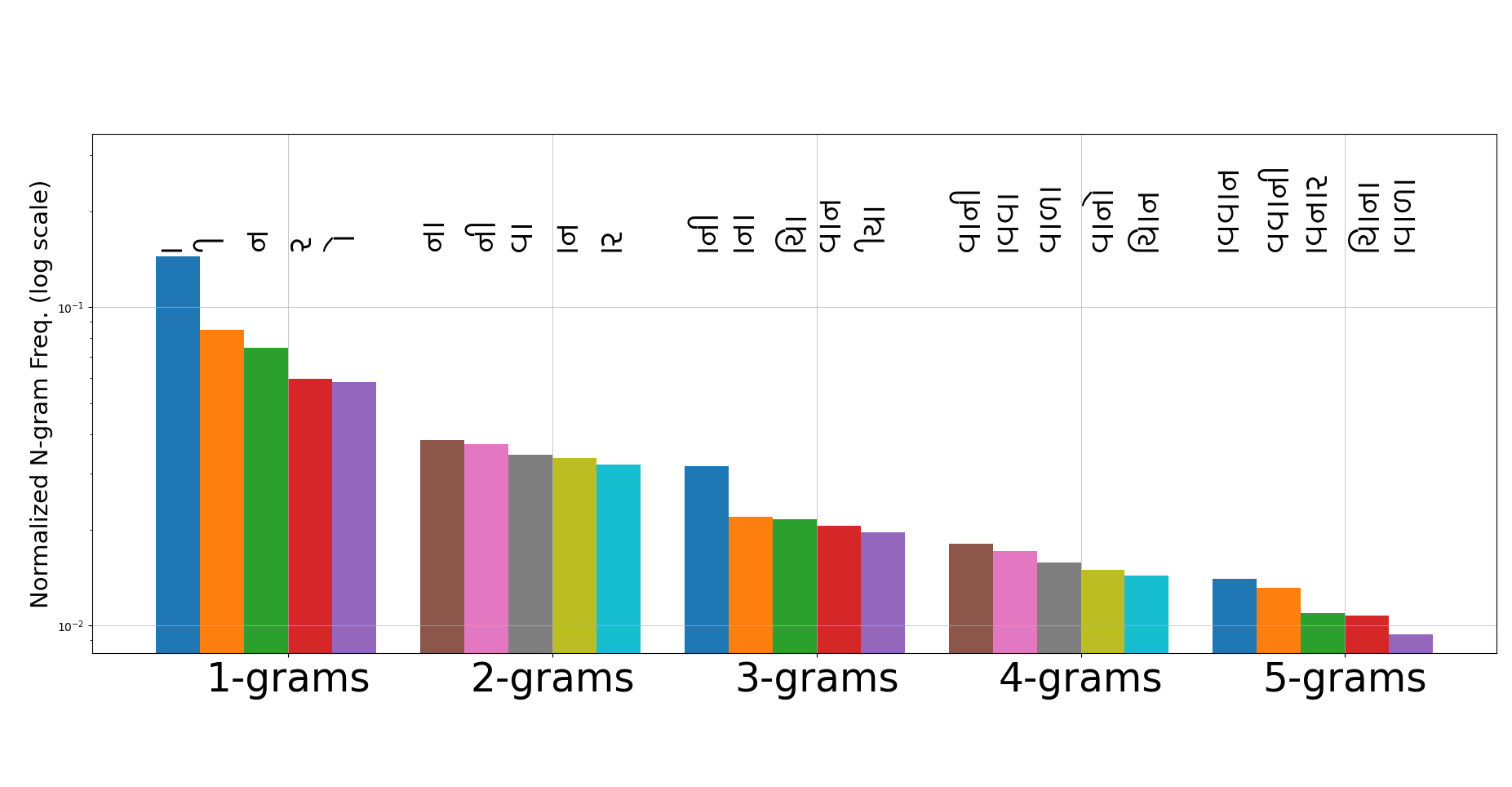}
    \end{subfigure}
    \begin{subfigure}[b]{0.435\textwidth}
    \centering
    \includegraphics[width=1\linewidth]{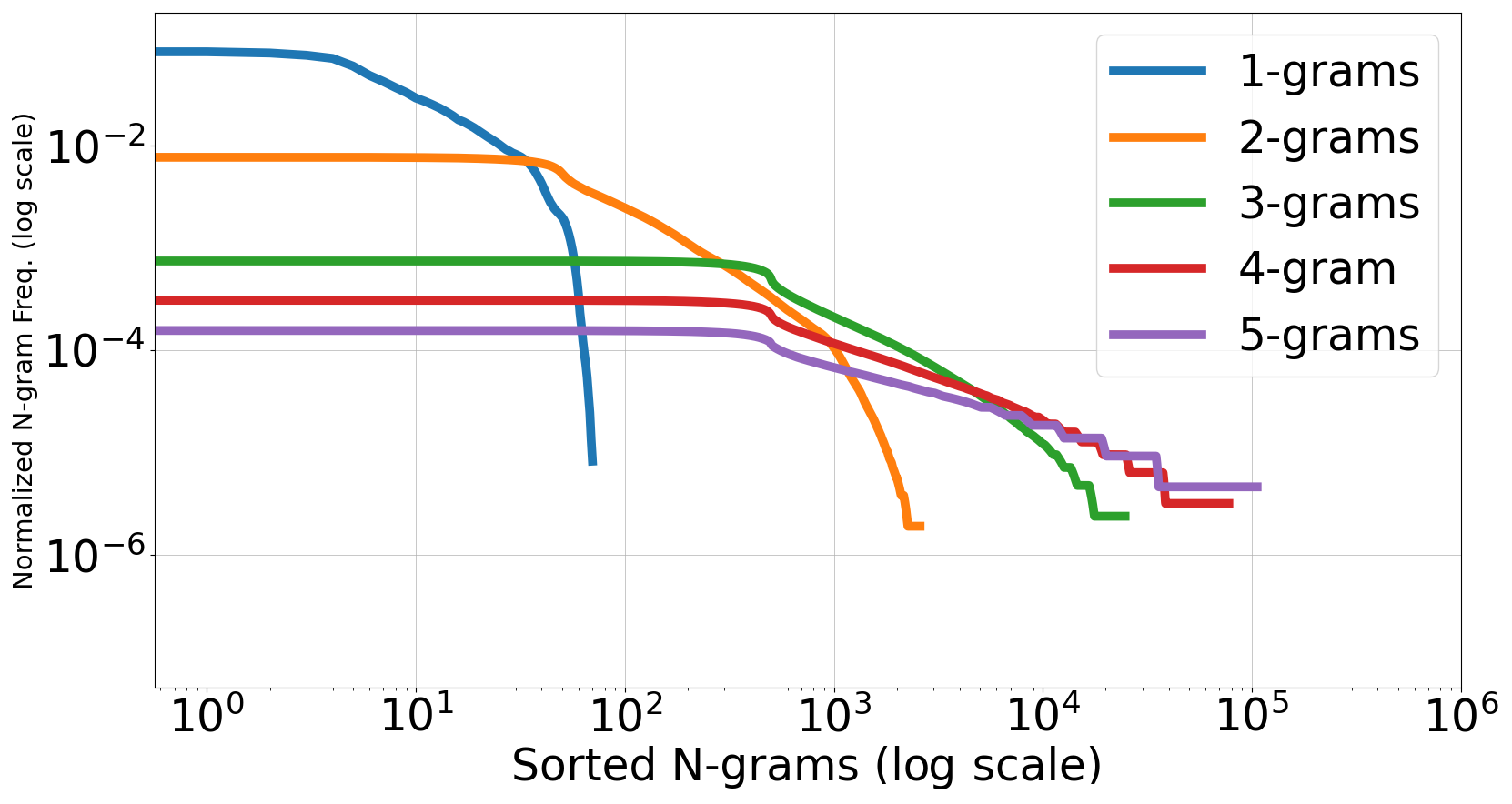}%
    \end{subfigure}
    \begin{subfigure}[b]{0.55\textwidth}
    \centering
    \includegraphics[trim=0 35 0 110,clip,width=1\linewidth]{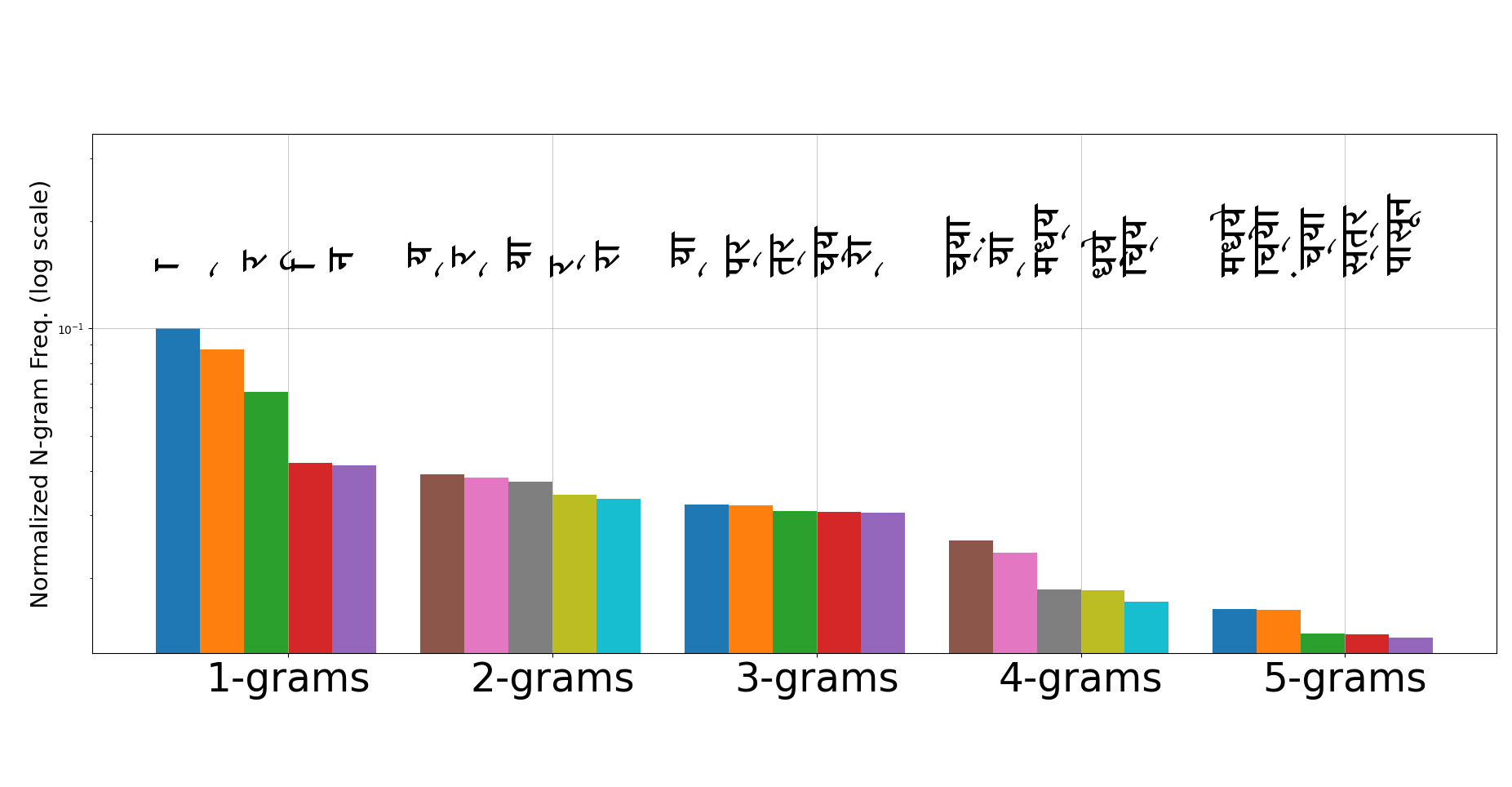}
    \end{subfigure}
    \begin{subfigure}[b]{0.435\textwidth}
    \centering
    \includegraphics[width=1\linewidth]{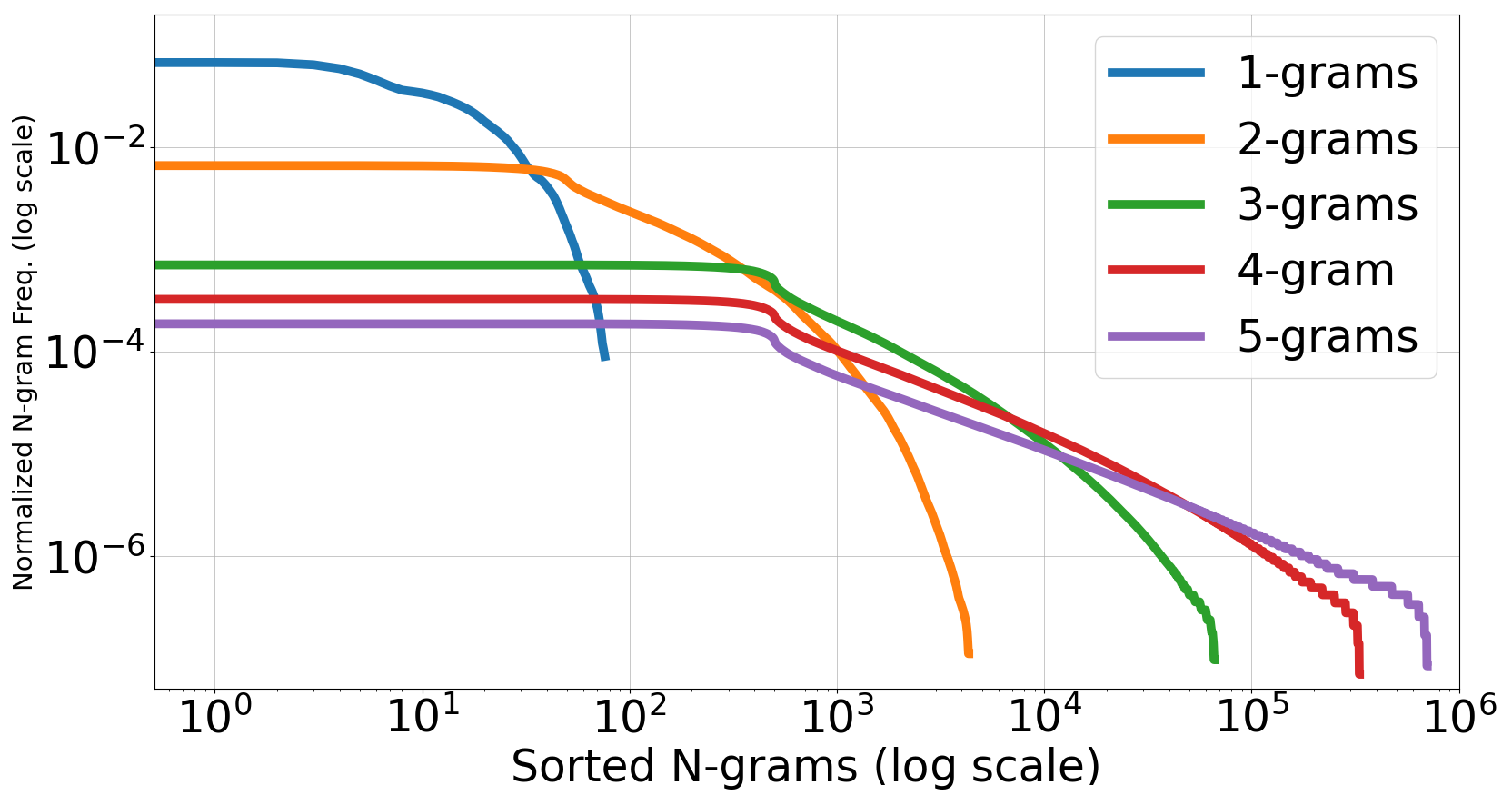}%
    \end{subfigure}
    \begin{subfigure}[b]{0.55\textwidth}
    \centering
    \includegraphics[trim=0 35 0 110,clip,width=1\linewidth]{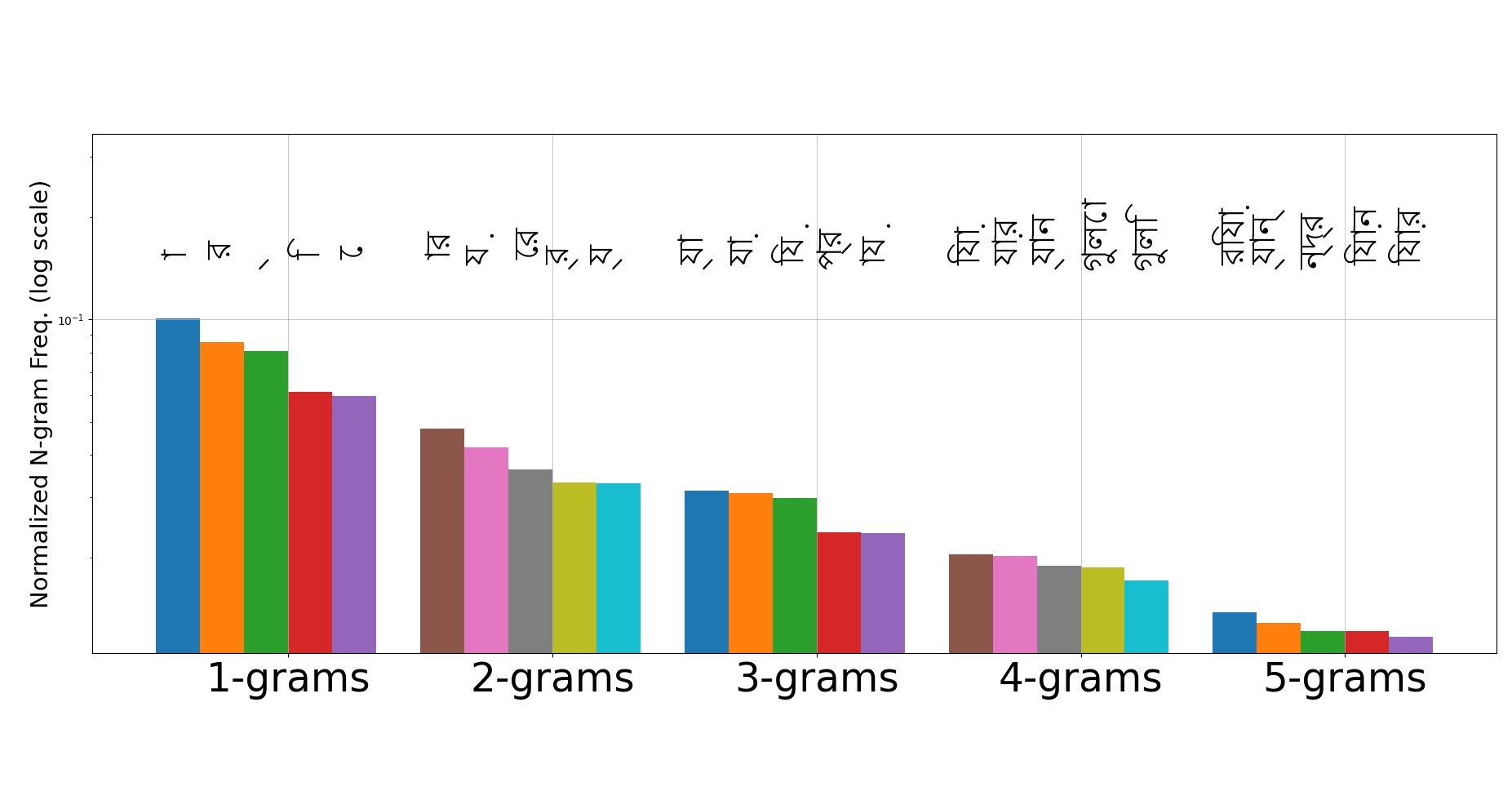}
    \end{subfigure}
    \begin{subfigure}[b]{0.435\textwidth}
    \centering
    \includegraphics[width=1\linewidth]{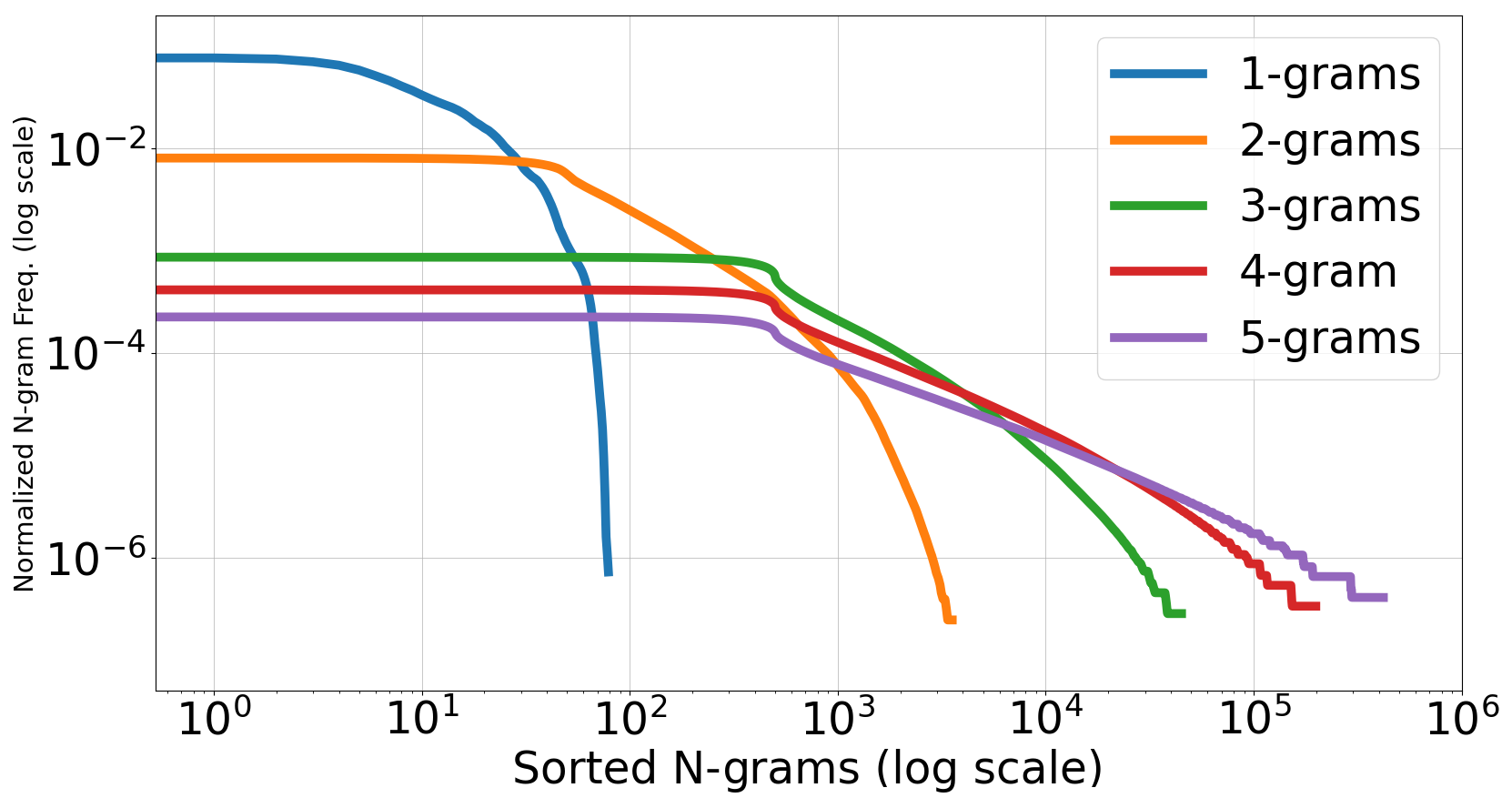}%
    \end{subfigure}
    \begin{subfigure}[b]{0.55\textwidth}
    \centering
    \includegraphics[trim=0 35 0 110,clip,width=1\linewidth]{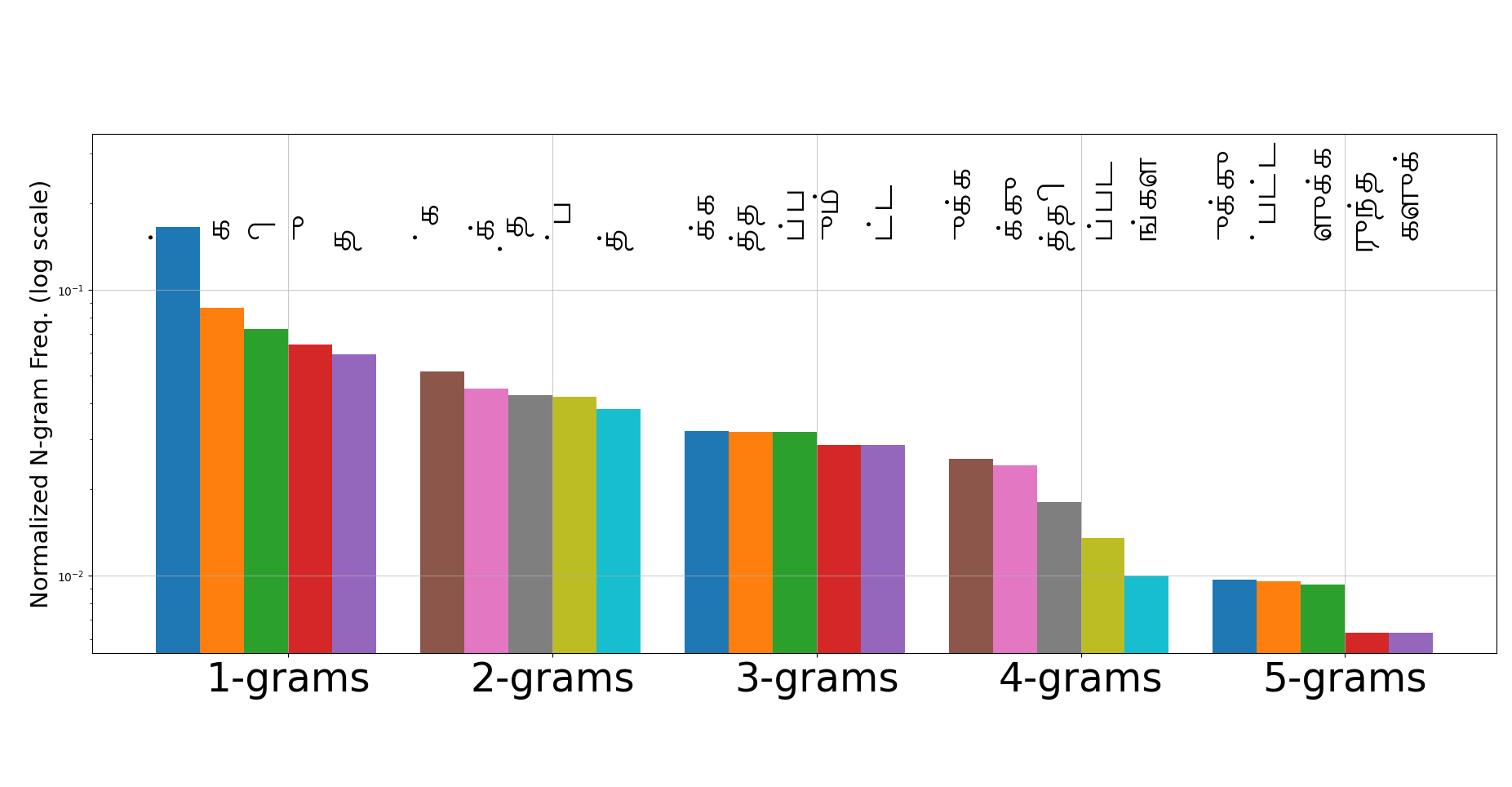}
    \end{subfigure}
    \begin{subfigure}[b]{0.435\textwidth}
    \centering
    \includegraphics[width=1\linewidth]{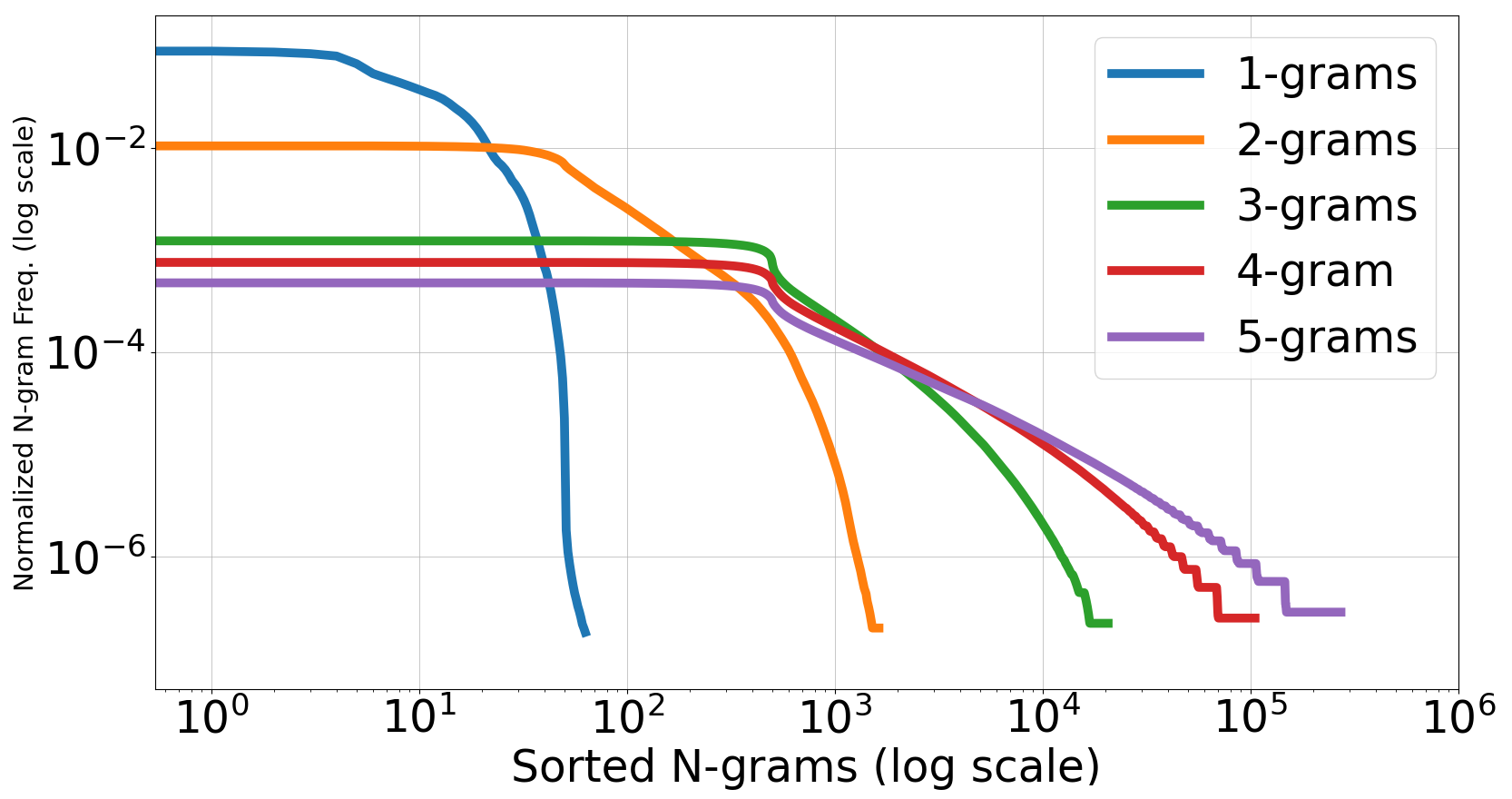}%
    \end{subfigure}
    \caption{Distribution of Char. n-grams ($n\in[1,5]$) from $2.5M$ words in English, Gujarati, Hindi, Bangla, and Tamil (top to bottom): Top-5 (left) and All (right).}
    \label{fig:ngramstats}
\end{figure}
Sample images of our synthetic data are shown in Fig.~\ref{fig:sample_synth_images}. For English, we use the models pre-trained on the $9M$ MJSynth and $8M$ SynthText images ~\cite{Jaderberg14c,gupta2016synthetic}. We generate $0.5M$ synthetic images in English with over $1200$ fonts for testing. As shown in Table~\ref{tab:synth_data}, English has a lower average word length than Indian languages. We list the Indian languages in the increasing order of language complexity, with visually similar scripts placed consecutively, in Table~\ref{tab:synth_data}. Gujarati is chosen as the entry point from English to Indian languages as it has the lowest word length among all Indian languages. Subsequently, like English, Gujarati does not have a top-connector line that connects different characters to form a word in Hindi and Bangla (refer to Fig.~\ref{fig:sample_scene_images} and~\ref{fig:sample_synth_images}). Also, the number of Unicode fonts available in Gujarati is fewer than those available in other Indian languages. Next, we choose Hindi, as Hindi characters are similar to Gujarati characters and the average word length of Hindi is higher than Gujarati. Bangla has comparable word length statistics with Hindi and shares the property of the top-connector line with Hindi. Still, we keep it after Hindi in the list as its characters are visually dissimilar and more complicated than Gujarati and Hindi. We use less than $100$ for fonts in Hindi, Bangla, and Telugu. We list Tamil after Bangla because these languages share similar vowels' appearance (see the glyphs above general characters in Fig.~\ref{fig:sample_synth_images}). Tamil and Malayalam have the highest variability in word length and visual complexity compared to other languages.  Please note that we have over $150$ fonts available in Tamil.

{\bf Real Datasets:} We also perform experiments on the real datasets from IIIT-ILST, MLT-17, and MLT-19 datasets (refer to Section~\ref{sec:relatedwork} for these datasets). To enlarge scene-text recognition research in complex and straight forward low-resource Indian Languages, we release $500$ and $2535$ annotated word images in Gujarati and Tamil. We crop the word images from $440$ annotated scene images, which we obtain by capturing and compiling Google images. We illustrate sample annotated images of different datasets in Fig.~\ref{fig:sample_scene_images}. Similar to MLT datasets, we annotate the Gujarati and Tamil datasets using four corner points around each word (see Tamil image at bottom-right of Fig.~\ref{fig:sample_scene_images}). IIIT-ILST dataset has two-point annotations leading to an issue of text from other words in the background of a cropped word image as shown in the Hindi scene at the top-middle of Fig.~\ref{fig:sample_scene_images}.

{\bf Motivation:} As discussed earlier in Section~\ref{sec:relatedwork}, most of the scene-text recognition works use the pre-trained Convolutional Neural Networks (CNN) layers for improving results. We now motivate the need for transfer learning of the complete recognition models discussed in Section~\ref{sec:Intro} and the models we use in Section~\ref{sec:models} among different languages. As discussed in these sections, the Recurrent  Neural  Networks  (RNNs) form another integral component of such reading models. Therefore, we illustrate the distribution of character-level n-grams they learn in Fig.~\ref{fig:ngramstats}\footnote{For plots on the right, we use moving average of 10, 100, 1000, 1000, 1000 for 1-grams, 2-grams, 3-grams, 4-grams, and 5-grams, respectively.} for the first five languages we discussed in the previous section (we notice that the last two languages also follow the similar trend). On the left, we show the frequency distribution of top-5 n-grams, ($n\in[1,5]$). On the right, we show the frequency distribution of all n-grams with $n\in[1,5]$. We use $2.5 M$ words from each language for these plots. We consider both capital and small letters separately for English, as it is crucial for the text recognition task. Despite this, we note that top-5 n-grams are composed of small letters. The Indian languages, however, do not have small and capital letters like English. However, the total number of English letters (given that small letters are different from capitals) is of the same order as Indian languages. The x-values ($<=100$) for the drops in 1-gram plots (blue curves) of Fig.~\ref{fig:ngramstats} also illustrates this. So it becomes possible to compare the distributions. Next, we note that most of the top-5 n-grams comprise vowels for all the languages. Moreover, the overall distributions are similar for all the languages. Hence, we propose that the RNN layers' transfer among the models of different languages is worth an investigation.

It is important to note the differences between the n-grams of English and Indian languages. Many of the top-5 n-grams in English are the complete word forms, which is not the case with Indian languages owing to their richness in inflections (or fusions)~\cite{vinithaerror}. Also, note that the second and the third 1-gram for Hindi and Bangla in Fig.~\ref{fig:ngramstats} (left), known as Halanta, is a common feature of top-5 Indic n-grams. The Halanta forms an essential part of joint glyphs or {\it aksharas} (as advocated by Vinitha et al.~\cite{vinithaerror}). 
In Figs.~\ref{fig:sample_scene_images} and~\ref{fig:sample_synth_images}, the vowels, or portions of the joint glyphs for word images in Indian languages, often appear above the top-connector line or below the generic consonants. All this, in addition to complex glyphs in Indian languages, makes transfer learning from English to Indian languages ineffective, which is detailed in Section~\ref{sec:results}. Thus, we also investigate the transferability of features among the Indic scene-text recognition models in the subsequent sections.
\section{Models}\label{sec:models}
This section explains the two models we use for transfer learning in Indian languages and a plug-in module we propose for learning the correction mechanism in the recognition systems.

\begin{figure}[t]
    \centering
    \includegraphics[width=0.975\linewidth]{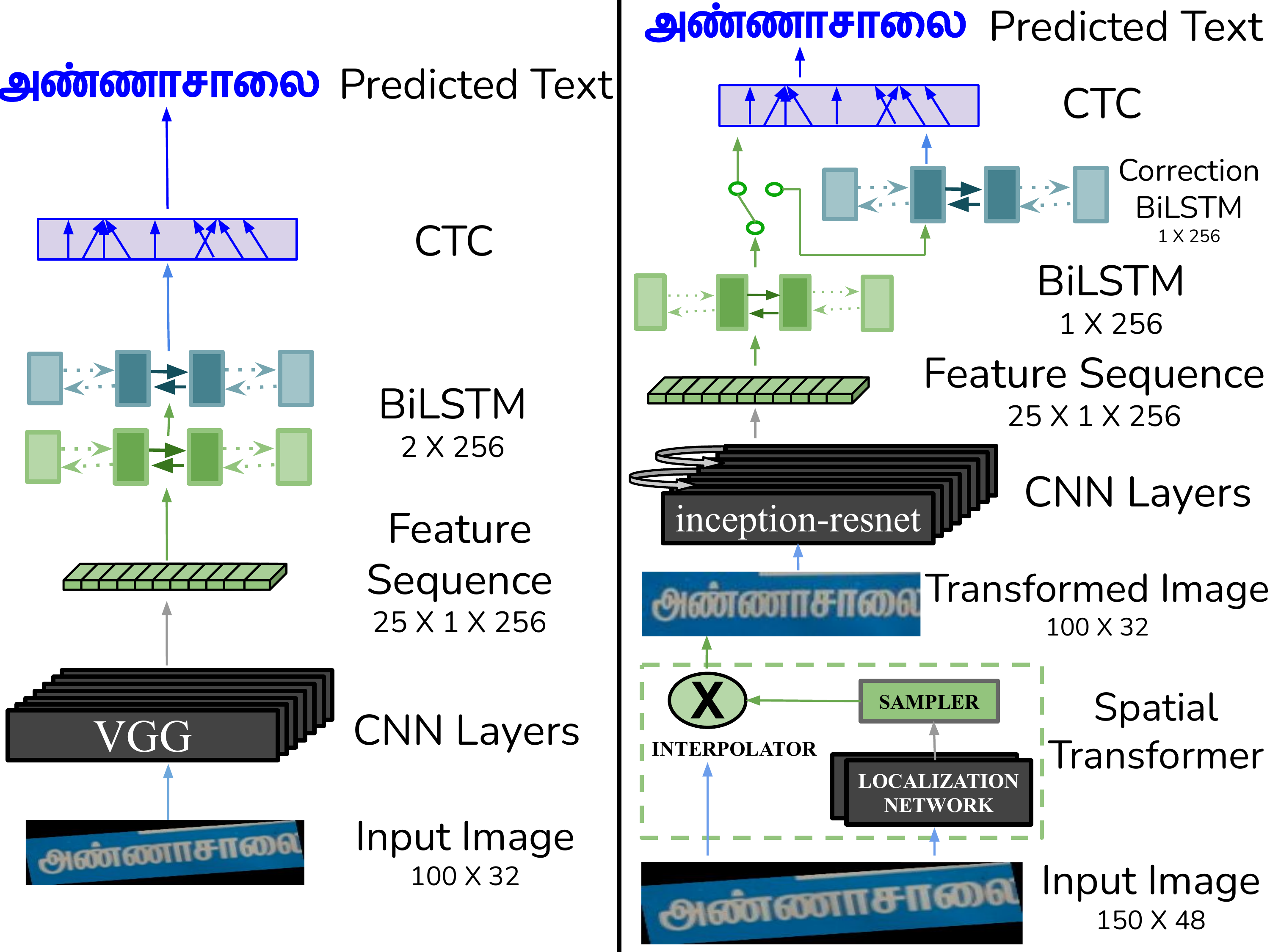}
    \caption{CRNN Model (left) and STAR-Net with a Correction BiLSTM (right).}
    \label{fig:models}
\end{figure}

{\bf CRNN Model:} The first model we train is Convolutional-Recurrent Neural Network (CRNN), which is the combination of CNN and RNN as shown in Fig.~\ref{fig:models} (left). The CRNN network architecture consists of three fundamental components, i) an encoder composed of the standard VGG model~\cite{simonyan2014very}, ii) a decoder consisting of RNN, and iii) a Connectionist Temporal Classification (CTC) layer to align the decoded sequence with ground truth. The CNN-based encoder consists of seven layers to extract feature representations from the input image. The model abandons fully connected layers for compactness and efficiency. It replaces standard squared pooling with $1\times2$ sized rectangular pooling windows for $3^{rd}$ and $4^{th}$ max-pooling layer to yield feature maps with a larger width. A two-layer Bi-directional Long Short-Term Memory (BiLSTM) model, each with a hidden size of $256$ units, then decodes the features. During the training phase, the CTC layer provides non-parameterized supervision to align the decoded predictions with the ground truth. The greedy decoding is used during the testing stage. We use the PyTorch implementation of the model by Shi et al.~\cite{shibY17end2end}.

{\bf STAR-Net:} As shown in Fig.~\ref{fig:models} (right), the STAR-Net model consists of three components, i) a Spatial Transformer to handle image distortions, ii) a Residue Feature Extractor consisting of a residue CNN and an RNN, and iii) a CTC layer to align the predicted and ground truth sequences. The transformer consists of a spatial attention mechanism achieved via a CNN-based localization network, a sample, and an interpolator. The localizer predicts the parameters of an affine transformation. The sampler and the nearest-neighbor interpolator use the transformation to obtain a better version of the input image. The transformed image acts as the input to the Residue Feature Extractor, which includes the CNN and a single-layer BiLSTM of $256$ units. The CNN used here is based on the inception-resnet architecture, which can extract robust image features required for the task of scene-text recognition~\cite{szegedy2017inception}. The CTC layer finally provides the non-parameterized supervision for text alignment. The overall model consists of $26$ convolutional layers and is end-to-end trainable~\cite{liu2016star}. 

{\bf Correction BiLSTM:} After training the STAR-Net model on a real dataset, we add a correction BiLSTM layer (of size $1\times256$), an end-to-end trainable module, to the end of the model (see Fig.~\ref{fig:models} top-right). We train the complete model again on the same dataset to implicitly learn the error correction mechanism.
\section{Experiments}\label{sec:experiments}
The images, resized to $150\times18$, form the input of STAR-Net. The spatial transformer module, as shown in Fig.~\ref{fig:models} (right), then outputs the image of size $100\times32$ . The inputs to the CNN Layers of CRNN and STAR-Net are of the same size, i.e., $100\times32$, and the output size is $25\times1\times256$. The STAR-Net localization network has four plain convolutional layers with $16$, $32$, $64$, and $128$ channels. Each layer has the filter size, stride, and padding size of $3$, $1$, and $1$, followed by a $2\times2$ max-pooling layer with a stride of $2$. Finally, a fully connected layer of size $256$ outputs the parameters which transform the input image. We train all our models on $2M$ or more synthetic word images as discussed in Section~\ref{sec:datasets}. We use the batch size of $16$ and the ADADELTA optimizer for stochastic gradient descent (SGD) for all the experiments~\cite{article}. The number of epochs varies between $10$ to $15$ for different experiments. We test our models on 0.5 M synthetic images for each language. We use the word images from IIIT-ILST, MLT-17, and MLT-19 datasets for testing on real datasets. 
We fine-tune the Bangla models on $1200$ training images and test them on $673$ validation images from the MLT-17 dataset to fairly compare with Bušta et al.~\cite{buvsta2017deep}. Similarly, we fine-tune only our best Hindi model on the MLT-19 dataset and test it on the IIIT-ILST dataset to compare with OCR-on-the-go (since it is also trained on real data)~\cite{saluja2019ocrgo}. To demonstrate generalizability, we also test our models on $3766$ Hindi images and $3691$ Bangla images available from MLT-19 datasets~\cite{nayef2019icdar2019}. For Gujarati and Tamil, we use $75\%$ of word images to fine-tune our models and the remaining $25\%$ for testing. 
\begin{table}[t]
\resizebox{0.8\textwidth}{!}
{%
\centering
\begin{tabular}{llrrrrrrrr}
\toprule
Language && CRNN-CRR && CRNN-WRR && STAR-Net-CRR && STAR-Net-WRR \\
\midrule
English && 77.13 && 38.21 && {\bf 86.04} && {\bf 57.28} \\
Gujarati && 94.43 && 81.85 && {\bf 97.80} && {\bf 91.40}\\
Hindi && 89.83 && 73.15 && {\bf 95.78} && {\bf 83.93}\\
Bangla && 91.54 && 70.76 && {\bf 95.52} && {\bf 82.79}\\
Tamil && 82.86 && 48.19 && {\bf 95.40} && {\bf 79.90}\\
Telugu && 87.31 && 58.01 && {\bf 92.54} && {\bf 71.97}\\
Malayalam && 92.12 && 70.56 && {\bf 95.84} && {\bf 82.10}\\
\bottomrule
\end{tabular}
}
\caption{Results of individual CRNN \& STAR-Net models on Synthetic Datasets.}
\label{tab:results_synthetic}
\end{table}
\section{Results}\label{sec:results}
In this section, we discuss the results of our experiments with i) individual models for each language, ii) the transfer learning from English to two Indian languages, and iii) the transfer learning from one Indian language to another.

{\bf Performance on Synthetic Datasets:} It is essential to compare the results on synthetic datasets of different languages sharing common backgrounds, as it provides a good intuition about the difficulty in reading different scripts. In Tables~\ref{tab:results_synthetic} and~\ref{tab:results_transferLearning_synth}, we present the results of our experiments with synthetic datasets.  As noted in Table~\ref{tab:results_synthetic}, the CRNN model achieves the Character Recognition Rates (CRRs) and Word Recognition Rates (WRRs) of i) 77.13\% and 38.21\% in English and ii) above $82\%$ and $48\%$ on the synthetic dataset of all the Indian languages (refer to columns 1 and 2 of Table~\ref{tab:results_synthetic}). The low accuracy on the English synthetic test set is due to the presence of more than 1200 different fonts (refer Section~\ref{sec:datasets}). Nevertheless, using a large number of fonts in training helps in generalizing the model for real settings~\cite{Jaderberg14c,gupta2016synthetic}. The STAR-Net achieves remarkably better performance than CRNN on all the datasets, with the CRRs and WRRs above $90.48$ and $65.02$ for Indian languages. The reason for this is spatial attention mechanism and powerful residual layers, as discussed in Section~\ref{sec:models}. As shown in columns 3 and 5 of Table~\ref{tab:results_synthetic}, the WRR of the models trained in Gujarati, Hindi, and Bangla are higher than the other three Indian languages despite common backgrounds. The experiments show that the scripts in latter languages pose a tougher reading challenge than the scripts in former languages. 
\begin{table}[t]
\resizebox{\textwidth}{!}
{%
\centering
\begin{tabular}{lrrrrrrrr}
\toprule
Language & & CRNN-CRR & & CRNN-WRR & & STAR-Net-CRR & & STAR-Net-WRR \\
\midrule
English $\rightarrow$ Gujarati & & 92.71 ({\bf 94.43}) & & 77.06 ({\bf 81.85}) & & 97.50 ({\bf  97.80}) & & 90.90 ({\bf 91.40}) \\
English $\rightarrow$ Hindi & &  88.11 ({\bf 89.83}) & & 70.12 ({\bf 73.15}) & & 94.50 ({\bf 95.78}) & & 80.90 ({\bf 83.93}) \\
\midrule
Gujarati  $\rightarrow$  Hindi & & {\bf 91.98} (89.83) & & 73.12 ({\bf 73.15}) & & {\bf 96.12} (95.78) & &  {\bf 84.32} (83.93)\\
Hindi $\rightarrow$ Bangla & & 91.13 ({\bf 91.54}) & & 70.22 ({\bf 70.76}) & & {\bf 95.66} (95.52) & & {\bf 82.81} (82.79) \\
Bangla $\rightarrow$ Tamil & & 81.18 ({\bf 82.86}) & & 44.74 ({\bf 48.19}) & & {\bf 95.95} (95.40) & & {\bf 81.73} (79.90) \\
Tamil $\rightarrow$ Telugu & & 87.20 ({\bf 87.31}) & &  56.24 ({\bf 58.01}) & & {\bf 93.25} (92.54) & & {\bf 74.04} (71.97) \\
Telugu $\rightarrow$ Malayalam & & 90.62 ({\bf 92.12}) & & 65.78 ({\bf 70.56}) & &  94.67 ({\bf 95.84}) & & 77.97 ({\bf 82.10}) \\
\bottomrule
\end{tabular}
}
\caption{Results of Transfer Learning (TL) on Synthetic Datasets. Parenthesis contain results from Table~\ref{tab:results_synthetic}. TL among Indic scripts improves STAR-Net results.}
\label{tab:results_transferLearning_synth}
\end{table}

We present the results of our transfer learning experiments on the synthetic datasets in Table~\ref{tab:results_transferLearning_synth}. The best individual model results from Table~\ref{tab:results_synthetic} are included in parenthesis for comparison. We begin with the English models as the base because the models have trained on over $1200$ fonts and $17M$ word images as discussed in Section~\ref{sec:datasets}, and are generic. However, in the first two rows of the table, we note that transferring the layers from the model trained on the English dataset to Gujarati and Hindi is inefficient in improving the results compared to the individual models. The possible reason for the inefficiency is that Indic scripts have many different visual and slightly different n-gram characteristics from English, as discussed in Section~\ref{sec:datasets}. We then note that as we try to apply transfer learning among Indian languages with CRNN (rows 3-7, columns 1-2 in Table~\ref{tab:results_transferLearning_synth}), only some combinations work well. However, with STAR-Net (rows 3-7, columns 3-4 in Table~\ref{tab:results_transferLearning_synth}), transfer learning helps improve results on the synthetic dataset from a simple language to a complex language\footnote{We also discovered experiments on transfer learning from a tricky language to a simple one to be effective but slightly lesser than the reported results.}. For Malayalam, we observe that the individual STAR-Net model is better than the one transferred from Telugu, perhaps due to high average word length (refer Section~\ref{sec:datasets}).
\begin{table}[H]
\resizebox{0.8\textwidth}{!}
{%
\centering
\begin{tabular}{lllcrr}
\toprule
Language & Dataset & \# Images & Model & CRR  & WRR \\
\midrule
&       &     & CRNN  & 84.93 & 72.08 \\
Gujarati &   ours    & 125 &    STAR-Net  & 88.55 & 74.60 \\
         & & & STAR-Net Eng$\rightarrow$Guj & 78.48 & 60.18\\
         & & & STAR-Net Hin$\rightarrow$Guj & {\bf 90.82} & {\bf 76.98} \\
\midrule
 &  &  & Mathew et al.~\cite{mathew2017benchmarking} &  75.60 & 42.90\\
 &  &  & CRNN & 78.84 & 46.56\\
& &  & STAR-Net  & 78.72 &  46.60\\
 Hindi & IIIT-ILST & 1150 & STAR-Net Eng$\rightarrow$Hin & 77.43 & 44.81 \\
 &  &  & STAR-Net Guj$\rightarrow$Hin & 79.12 & 47.79\\
 &  &  & OCR-on-the-go~\cite{saluja2019ocrgo} & - & 51.09 \\
 &  &  & STAR-Net Guj$\rightarrow$Hin FT\footnotemark{} & {\bf 83.64} & {\bf 56.77} \\
\midrule
 &  &  & CRNN  & 86.56 & 64.97\\
Hindi & MLT-19 & 3766 & STAR-Net  & 86.53 &  65.79\\
 & & & STAR-Net Guj$\rightarrow$Hin & {\bf 89.42} & {\bf 72.96} \\
\midrule
 &  &  & Bušta et al.~\cite{buvsta2018e2e} &   68.60 & 34.20\\
Bangla & MLT-17 & 673 & CRNN  & 71.16 & 52.74 \\
 &  &  & STAR-Net  & 71.56 & 55.48 \\ 
 & & & STAR-Net Hin$\rightarrow$Ban & 72.16 & 57.01 \\
 & & & W/t Correction BiLSTM & {\bf 83.30} & {\bf 58.07} \\
\midrule
&  & & CRNN  &  81.93 & 74.26 \\
Bangla & MLT-19 & 3691 & STAR-Net  &  82.80 & 77.48\\
 &  &  & STAR-Net Hin$\rightarrow$Ban & {\bf 82.91} & {\bf 78.02} \\
\midrule

 &      &  & CRNN  &  {\bf 90.17} & 70.44\\
 Tamil & ours  & 634 &  STAR-Net &  89.69 & 71.54\\ 
 &   &      & STAR-Net Ban$\rightarrow$Tam & 89.97 & {\bf 72.95}\\
\midrule
 &  &  & Mathew et al.~\cite{mathew2017benchmarking} &  {\bf 86.20} & 57.20\\
Telugu & IIIT-ILST & 1211 & CRNN  & 81.91 & 58.13\\
 &  &  & STAR-Net & 82.21 & 59.12\\
 &  &  & STAR-Net Tam$\rightarrow$Tel & 82.39 & {\bf 62.13} \\
\midrule
 &  &  & Mathew et al.~\cite{mathew2017benchmarking} & {\bf 92.80}  & 73.40\\
Malayalam & IIIT-ILST & 807 & CRNN  & 84.12 & 70.36 \\ 
 &  &  & STAR-Net  & 91.50 & 72.73 \\ 
 & & & STAR-Net Tel$\rightarrow$Mal & 92.70 & {\bf 75.21}\\
\bottomrule
\end{tabular}
}
\caption{Results on Real Datasets. FT indicates Fine-Tuned.}
\label{tab:results_real}
\end{table} \footnotetext{Fine-tuned on MLT-19 dataset as discussed earlier. We fine-tune all the layers.}
\begin{figure}[h!t]
    \centering
    \includegraphics[trim=20 232 0 20,width=\linewidth]{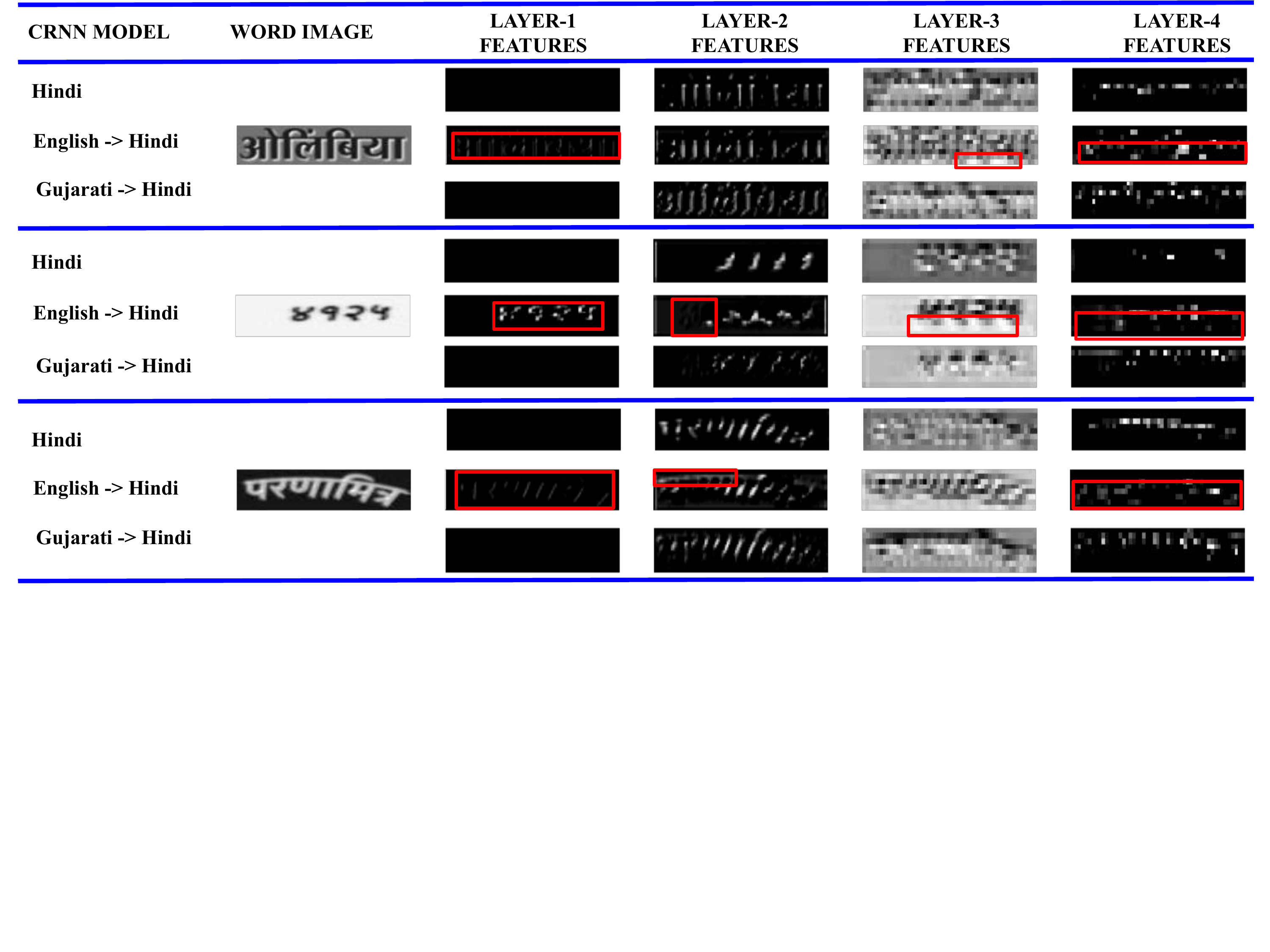}\\%
    \includegraphics[trim=20 250 0 20,width=\linewidth]{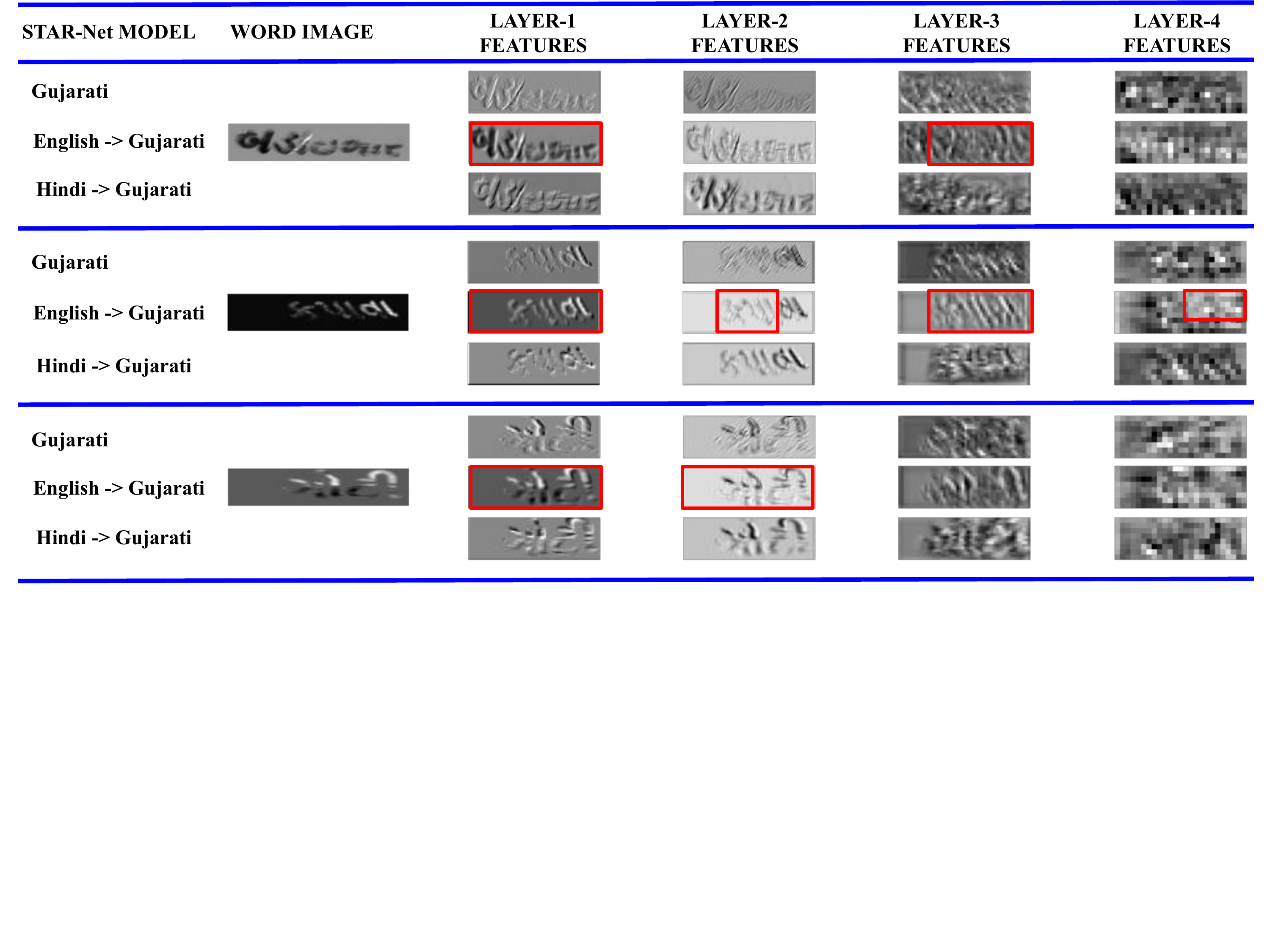}%
    \caption{CNN Layers visualization in the {\bf Top:} CRNN models trained on Hindi, English$\rightarrow$Hindi, and Gujarati$\rightarrow$Hindi; and {\bf Bottom:} STAR-Net models trained on Gujarati, English$\rightarrow$Gujarati, and Hindi$\rightarrow$Gujarati. Red boxes indicate the regions where the features for the model transferred from English are activated (as white), whereas the features from the other two models are not.}
    \label{fig:visualizations}
\end{figure}
{\bf Performance on Real Datasets:} Table~\ref{tab:results_real} depicts the performance of our models on the real datasets. At first, we observe that for each Indian language, the overall performance of the individual STAR-Net model is better than the individual CRNN model (except for Gujarati and Hindi, where the results are very close). Based on this and similar observations in the previous section, we present the results of transfer learning experiments on real datasets only with the STAR-Net model\footnote{We also tried transfer learning with CRNN; STAR-Net was more effective.}. Next, similar to the previous section, we observe that the transfer learning from English to Gujarati and Hindi IIIT-ILST datasets (rows $3$ and $8$ in Table~\ref{tab:results_real}) is not as effective as individual models in these Indian languages (rows $2$ and $7$ in Table~\ref{tab:results_real}). Finally, we observe that the performance improves with the transfer learning from a simple language to a complex language, except for Hindi$\rightarrow$Gujarati, for which Hindi is the only most straightforward choice. We achieve performance better than the previous works, i.e., Bušta et al.~\cite{buvsta2017deep}, Mathew et al. ~\cite{mathew2017benchmarking}, and OCR-on-the-go~\cite{saluja2019ocrgo}. Overall, we observe the increase in WRRs by $6\%$, $5\%$, $2\%$ and $23\%$ on IIIT-ILST Hindi, Telugu, and Malayalam, and MLT-17 Bangla datasets compared to the previous works. On the MLT-19 Hindi and Bangla datasets, we achieve gains of $8\%$ and $4\%$ in WRR over the baseline individual CRNN models. On the datasets we release for Gujarati and Tamil, we improve the baselines by $5\%$ and $3\%$ increase in WRRs. We present the qualitative results of our baseline CRNN models as well as best transfer learning models in Fig.~\ref{fig:sample_scene_images}. The green and red colors represent the correct predictions and errors, respectively. ``\_" represents the missing character. As can be seen, most of the mistakes are single-character errors.

Since we observe the highest gain of $23\%$ in WRR (and $4\%$ in CRR) for the MLT-17 Bangla dataset (Table~\ref{tab:results_real}), we further try to improve these results. We plug in in the correction BiLSTM (refer Section~\ref{sec:models}) to the best model (row 18 of Table~\ref{tab:results_real}). 
The results are shown in row 19 of Table~\ref{tab:results_real}. As shown, the correction BiLSTM improves the CRR further by a notable margin of $11\%$ since the BiLSTM works on character level. We also observe the $1\%$ WRR gain, thereby achieving the overall $24\%$ WRR gain (and $15\%$ CRR gain) over Bušta et al.~\cite{buvsta2017deep}.

{\bf Features Visualization:} In Fig.~\ref{fig:visualizations} for the CRNN model (top three triplets), we visualize the learned CNN layers of the individual Hindi model, the ``English  $\rightarrow$Hindi" model, and the ``Gujarati$\rightarrow$Hindi" model. The red boxes are the regions where the first four CNN layers of the model transferred from English to Hindi are different from the other two models. The feature visualization again strengthens our claim that transfer from the English reading model to any Indian language dataset is inefficient. We notice a similar trend for the Gujarati STAR-Net models, though the initial CNN layers look very similar to word images (bottom three triplets in Fig.~\ref{fig:visualizations}). The similarity also demonstrates the better learnability of STAR-Net compared to CRNN, as observed in previous sections.
\section{Conclusion}~\label{sec:conclusions}
We generated $2.5M$ or more synthetic images in six different Indian languages with varying complexities to investigate the language transfers for two scene-text recognition models. The underlying view is that the transfer of image features is standard in deep models, and the transfer of language text features is a plausible and natural choice for the reading models. However, we observe that transferring the generic English photo-OCR models (trained on over $1200$ fonts) to Indian languages is inefficient. Our models transferred from one Indian language to another perform better than the previous works or the new baselines we created for individual languages. We, therefore, set the new benchmarks for scene-text recognition in low-resource Indian languages. The proposed Correction BiLSTM, when plugged into the STAR-Net model and trained end-to-end, further improves the results. 
\bibliographystyle{splncs04}
\bibliography{paper}
\end{document}